\documentclass{article}

\usepackage[preprint,nonatbib]{neurips_2026}

\usepackage[utf8]{inputenc}
\usepackage[T1]{fontenc}
\usepackage{url}
\usepackage{booktabs}
\usepackage{amsmath}
\usepackage{amssymb}
\usepackage{graphicx}
\usepackage{float}
\usepackage{wrapfig}
\usepackage{microtype}
\usepackage[table,dvipsnames]{xcolor}
\usepackage{array}
\usepackage{tabularx}
\usepackage{ragged2e}
\usepackage{pifont}
\usepackage{enumitem}
\usepackage[most]{tcolorbox}

\definecolor{ReportInk}{HTML}{2B2540}
\definecolor{ReportTeal}{HTML}{7C3AED}
\definecolor{ReportBlue}{HTML}{6D28D9}
\definecolor{ReportOrange}{HTML}{C026D3}
\definecolor{ReportLine}{HTML}{E3DCF4}
\definecolor{PromptBack}{HTML}{FBFAFF}
\newcolumntype{Y}{>{\RaggedRight\arraybackslash}X}
\newcolumntype{Z}{>{\RaggedLeft\arraybackslash}X}
\newcommand{\tablehead}{\rowcolor{ReportTeal!10}}
\newcommand{\cmark}{\ding{51}}
\newcommand{\xmark}{\ding{55}}

\newtcblisting{promptlisting}[1][]{%
  enhanced,
  breakable,
  listing only,
  listing engine=listings,
  colback=PromptBack,
  colframe=ReportLine,
  coltitle=ReportInk,
  fonttitle=\bfseries\footnotesize,
  boxrule=0.45pt,
  arc=1.2mm,
  left=2mm,
  right=2mm,
  top=1.1mm,
  bottom=1.1mm,
  borderline west={1.2pt}{0pt}{ReportTeal!70},
  listing options={
    basicstyle=\ttfamily\scriptsize,
    breaklines=true,
    columns=fullflexible,
    keepspaces=true,
    showstringspaces=false
  },
  #1
}

\usepackage[backend=biber,style=numeric-comp,sorting=none,maxbibnames=6,giveninits=true]{biblatex}
\newcommand{\citet}[1]{\textcite{#1}}
\newcommand{\citep}[1]{\parencite{#1}}

\usepackage{hyperref}

\title{StructAgent: Harness Long-horizon Digital Agents with Unified Causal Structure}
\author{%
Wenyi Wu$^{1,2,\S}$,
Sibo Zhu$^{1,2,\S}$,
Kun Zhou$^{2,\dagger}$,
Aayush Salvi$^{1}$, Zixuan Song$^{1}$, Biwei Huang$^{1,2}$\\
$^{1}$University of California, San Diego \quad
$^{2}$Aether AI Lab\\
{\footnotesize \textsuperscript{\S}Work done during an internship at Aether AI.}\\
{\footnotesize $^\dagger$Project Lead \& Corresponding Author: \texttt{franciskunzhou@gmail.com}.}\\
}

\begin{document}
\maketitle

\begin{abstract}
Recent advances in large language models (LLMs) and vision-language models (VLMs) have enabled increasingly capable digital agents for computer use. However, real-world tasks are often long-horizon and involve evolving contexts containing accumulated observations, intermediate edits, failed attempts, and partially completed executions. Existing agents typically operate over raw interaction history, making task progress difficult to interpret, verify, and recover, which ultimately limits reliable long-horizon execution. In this paper, we argue that addressing this challenge requires explicitly structuring both the agent's state and workflow around a unified causal representation of task progress. We present \textbf{StructAgent}, a state-centered framework that introduces a unified state for maintaining compact, verifiable task progress and a structured workflow that regulates progress through verifier-backed state transitions. Building on this design, StructAgent further enables explicit progress checkpointing, evidence-driven task completion, targeted failure recovery, and tool-supported execution, while ensuring that all progress updates remain grounded in verification. Extensive experiments demonstrate that StructAgent consistently improves a wide range of LLM and VLM backbones on long-horizon computer-use tasks. On OSWorld-Verified, it improves Qwen3.5-9B from 27.0\% to 46.9\% success rate and Qwen3.5-27B from 31.6\% to 62.2\%, while achieving a new open-source state of the art of 78.9\% with MiniMax-M3. Moreover, the same framework generalizes beyond desktop environments to Minecraft, demonstrating the generality of our design.
\end{abstract}

\begin{center}
{\normalsize
\raisebox{-0.16em}{\includegraphics[height=1.5em]{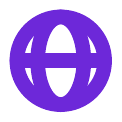}}\,\href{https://wenyiwu0111.github.io/structagent-page/}{\textbf{Project Page}}%
\hspace{2.4em}%
\raisebox{-0.16em}{\includegraphics[height=1.5em]{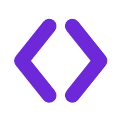}}\,\href{https://github.com/WenyiWU0111/StructAgent}{\textbf{Code}}%
}
\end{center}

\begin{center}
\emph{``The purpose of abstracting is not to be vague, but to create a new semantic level in which one can be absolutely precise.''}
--- Edsger W. Dijkstra
\end{center}

\section{Introduction}
\label{sec:intro}

\begin{figure}[t]
  \centering
  \includegraphics[width=\linewidth]{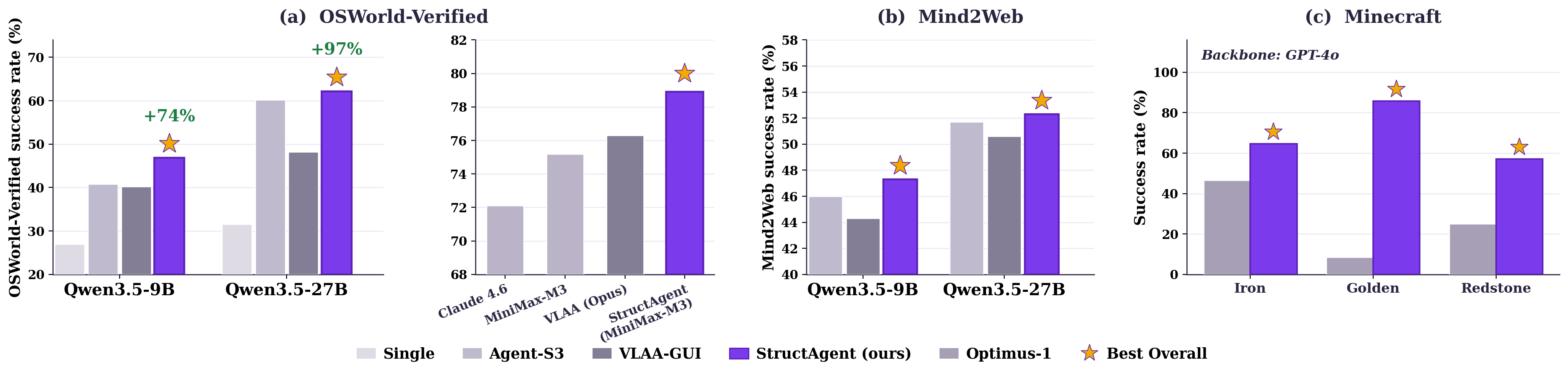}
  \caption{Results on OSWorld-Verified, Mind2Web, and
Minecraft. StructAgent improves matched open backbones and reaches
\textbf{78.9\%} with MiniMax-M3, the state-of-the-art open-source model result.}

  \label{fig:osworld_intro_result}
\end{figure}
\vspace{-0.2em}

Driven by scaling laws, large language models (LLMs) and vision-language models (VLMs) have become increasingly capable, demonstrating stronger abilities in perception, reasoning, and planning~\cite{zhao2023survey, wang2023surveyagents}. Building on these advances, digital agents have emerged as a promising paradigm for automating user tasks in computer environments~\cite{xie2024osworld,xlang2025osworldverified}. It completes user requests by observing the digital environment through visual screenshots or accessibility information, taking actions such as clicking, typing, and invoking system tools when necessary~\cite{tan2024cradle,wu2024oscopilot,agents2_2025,han2026vlaagui}.

In real-world applications, however, digital agents must execute long-horizon, complex workflows involving information retrieval, document editing, cross-application coordination, while preserving intermediate results across many steps. This setting is particularly challenging because the agent must continually manage an evolving context that accumulates large amounts of factual information, intermediate edits, failed attempts, and partially completed executions. As this context grows, critical information can become buried and the current task state increasingly ambiguous. Consequently, the agent struggles to maintain a clear causal understanding of task progress, making it difficult to promptly detect abnormal states, identify intermediate errors, and perform accurate planning and execution~\cite{yang2026ossymphony,han2026vlaagui}. 

To solve it, the agent’s working process must be explicitly organized to support tracing the effects of individual decisions, identifying the key factors that drive task progress, and retaining the minimum sufficient context for future execution. Therefore, we focus on two fundamental elements of long-horizon agents: \textbf{State} and \textbf{Workflow}. Our goal is to endow both with a unified, minimum-sufficient causal structure that provides a common foundation for planning, acting, and verification. To this end, we introduce two complementary components: a unified state, which provides all modules with a shared interface for reading and writing, and a structured workflow, which governs how progress is propagated across components through transparent state transitions. Concretely, the unified state consists of current requirements, useful values, and verified evidences, enabling all modules to access a consistent and verifiable representation of task progress. The structured workflow establishes a fixed execution loop that iteratively invokes the planner, actor, and verifier to drive state updates, thereby preserving a transparent causal chain throughout long-horizon execution.

Building on this design, we introduce \textbf{StructAgent}, a state-centered framework for long-horizon computer use. Beyond organizing planning, acting, and verification into a unified control loop, owing to the unified state and structured workflow designs, StructAgent enables explicit progress checkpointing through verifier-backed state updates, evidence-driven task completion instead of self-reported \textsc{Done} claims, targeted failure recovery based on verified evidences, and tool-supported execution that assists execution without bypassing state verification. Together, these mechanisms make long-horizon execution transparent, controllable, and robust.

Empirically, our agent framework is general to improve different LLM and VLM backbones on computer use tasks. On OSWorld-Verified, StructAgent improves Qwen3.5-9B from 27.0\% to 46.9\% success rate and Qwen3.5-27B from 31.6\% to 62.2\%. Using MiniMax-M3 backbone~\cite{lai2026minimaxsparse}, StructAgent reaches 78.9\%, achieving a new state-of-the-art performance among open-source methods on OSWorld-Verified~\cite{xlang2025osworldverified}, as summarized in Figure~\ref{fig:osworld_intro_result}. 
A Minecraft instantiation shows that our agent framework can be applied beyond desktop GUI benchmarks and lead to significant improvement, by replacing desktop evidence sources with inventory-based verification.

We summarize our contributions as follows:
\begin{itemize}[leftmargin=0pt,label={},itemsep=0.18em,topsep=0.2em]
\item \textbf{1. Unified state.}
We introduce a compact, typed, and auditable state that provides planning, acting, and verification with a shared representation of task progress.

\item \textbf{2. Structured workflow.}
We design a planning-acting-verification loop in which only verifier-backed decisions can commit, preserve, or invalidate progress, while memory, tools, and recovery provide state-conditioned support.

\item \textbf{3. Broad empirical validation.}
StructAgent achieves a new open-model state-of-the-art on OSWorld with MiniMax-M3, and generalizes the same structured state to web and Minecraft environments.
\end{itemize}

\section{Preliminaries}
\label{sec:preliminaries}

\paragraph{Digital Agent Task.}
A digital agent aims to automate computer-use tasks by interacting with a digital environment to fulfill a user instruction $u$. At each interaction step $t$, the agent receives an observation $o_t$, which may consist of a screenshot, accessibility information, or tool feedback, and selects the next action according to
\begin{equation}
a_t \sim \pi_\theta(\cdot \mid u, o_t), \qquad
o_{t+1} = \mathcal{E}(o_t, a_t),
\label{eq:cua-interaction}
\end{equation}
where $a_t \in \mathcal{A}$, and the action space $\mathcal{A}$ includes GUI operations (e.g., clicking, typing, scrolling, and waiting) as well as available system and application tools. Here, $\pi_\theta$ denotes the agent policy, and $\mathcal{E}$ is the environment transition induced by executing $a_t$. The interaction continues until the environment verifies that the execution satisfies the user instruction, at which point the task is considered successful.


\paragraph{Agent Context Management.}
Before $t$-th turn, the agent accumulates a raw interaction context $[o_1,a_1,\cdots, o_t,a_t]$, together with intermediate outputs, tool feedback, and environment responses. As the trajectory grows, the context increasingly mixes task-relevant facts with failed attempts, transient observations, intermediate edits, and unsupported completion claims. Effective long-horizon execution therefore requires the agent to distinguish past events from the information that remains valid and useful for future decisions. StructAgent addresses this by re-organizing the raw interaction history into an internal state $s_t$, which retains the compact, persistent, and verifiable information for planning, acting, and verification.


\begin{figure}[t]
  \centering
  \includegraphics[width=\linewidth]{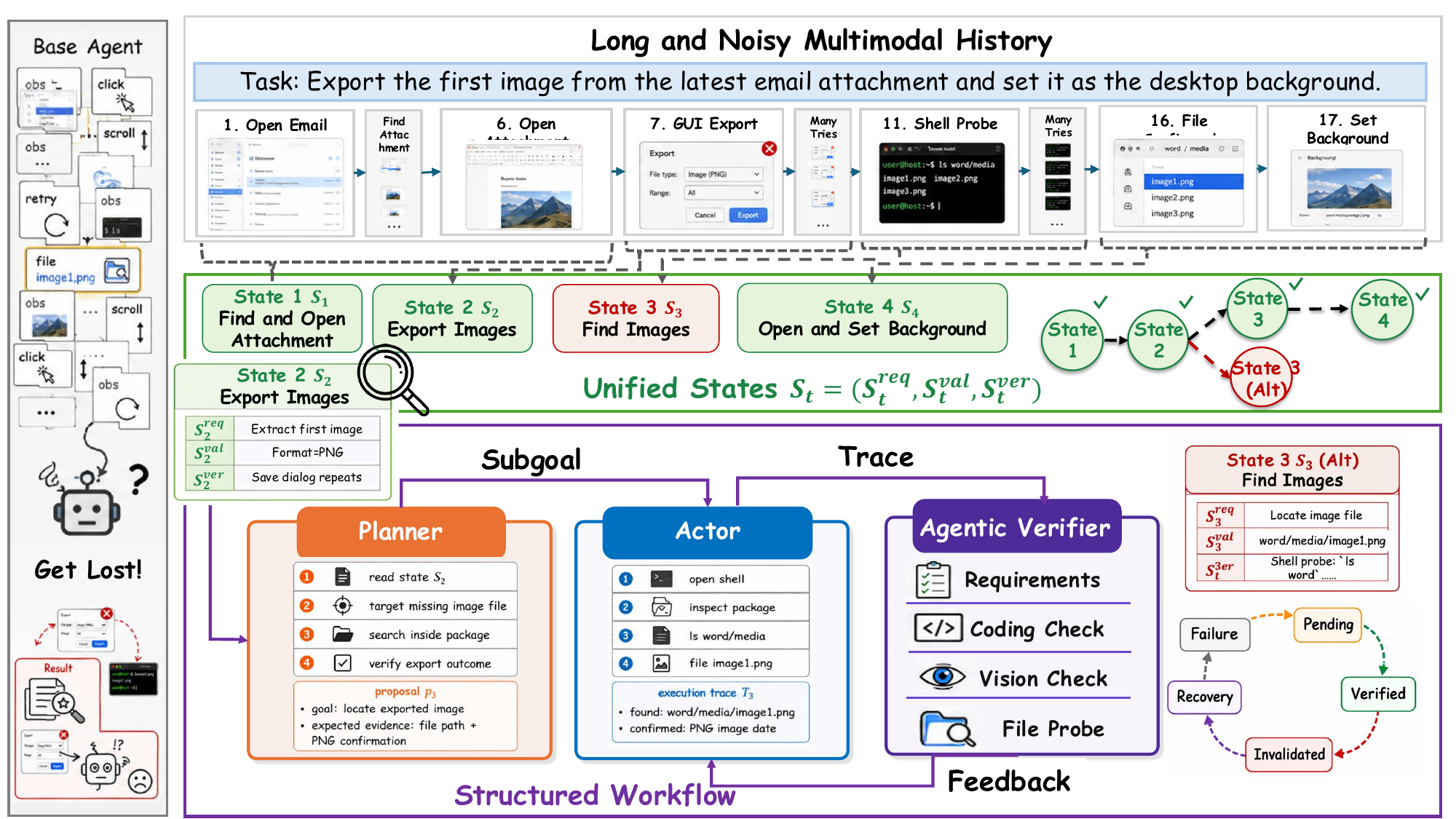}
  \caption{StructAgent overview. A unified state provides the shared progress
interface, while a structured workflow controls how planning, acting, and
verification interact with it. Planner and actor outputs can propose progress,
but only verifier-backed decisions commit or invalidate state.}
  \label{fig:overview}
\end{figure}

\section{StructAgent}
\label{sec:method}
To better support digital agents in long-horizon tasks, we propose StructAgent, a general framework for structured agent context management. StructAgent introduces a unified causal structure composed of a unified state and a structured workflow. Together, these components enable more transparent and controllable planning and execution, while naturally supporting advanced capabilities such as progress checkpointing and targeted failure recovery.



\subsection{Unified agent state}
\label{sec:state}

The unified agent state aims to turn a long and noisy interaction history into a compact state representation \(s_t\) that planning, acting, and verification can share. 
This makes \(s_t\) both the shared interface for reading task progress and the explicit causal surface through which progress can be updated.

\paragraph{State schema.}
Instead of asking each module to infer progress from raw trajectories, we maintain a state \(s_t\) that records what the current task still requires, what information should be carried forward, and what evidence supports the current progress:
\begin{equation}
s_t =
\left(
s_t^{\mathrm{req}},
s_t^{\mathrm{val}},
s_t^{\mathrm{ver}}
\right),
\label{eq:state}
\end{equation}
where \(s_t^{\mathrm{req}}\) stores current requirements,
\(s_t^{\mathrm{val}}\) stores useful values, and
\(s_t^{\mathrm{ver}}\) stores verified evidences. 
In this way, $s_t$ retains only the minimum sufficient information required to guide future planning and execution.
Concretely, \(s_t^{\mathrm{req}}\) stores the requirements of the current subgoal, specifying what must be achieved and what the verifier should check. \(s_t^{\mathrm{val}}\) stores useful values discovered during execution, such as file paths, URLs, selected entities, and extracted fields.
\(s_t^{\mathrm{ver}}\) stores verified evidences collected by the verifier during the agentic probing process, so later turns can audit and reuse this information.

\paragraph{Causality-oriented state transition.}
To ensure that the state $s_t$ faithfully reflects the true task progress, we design a causality-oriented state transition mechanism. In StructAgent, every state transition must be triggered by an explicit verifier-backed decision. Consequently, each progress update has a clear causal dependency on verifiable evidence, making the evolution of the state transparent and auditable.
Concretely, each requirement $r_i \in s_t^{\mathrm{req}}$ is associated with a status $\sigma_{i,t} \in \{\textsc{Pending},\ \textsc{Verified},\ \textsc{Invalidated}\}$, which respectively indicates that the requirement is not confirmed, verified with sufficient verified evidences, or invalidated by conflicting verified evidences. After the actor completes a subgoal, the verifier examines the task-relevant evidence and produces a verification decision $d_t$, determining whether each affected requirement should be verified, invalidated, or left unchanged.
The verification decision then drives the state transition according to
\begin{equation}
\small
\sigma_{i,t+1} =
\begin{cases}
\textsc{Verified}, & d_t \text{ verifies } r_i,\\
\textsc{Invalidated}, & d_t \text{ invalidates } r_i,\\
\sigma_{i,t}, & \text{otherwise}.
\end{cases}
\label{eq:requirement-update}
\end{equation}
It ensures that only verified evidences can advance or revoke task progress. If the verifier additionally identifies useful values required by subsequent execution, such as file paths, URLs, and selected entities, they are written into $s_{t+1}^{\mathrm{val}}$ together with their supporting verified evidences in $s_{t+1}^{\mathrm{ver}}$.

\subsection{Structured agent workflow}
\label{sec:workflow}

Using the current state \(s_t\), StructAgent runs a
planning-acting-verification workflow:
\begin{equation}
s_t
\xrightarrow{\mathrm{Planner}}
g_t
\xrightarrow{\mathrm{Actor}}
\tau_t
\xrightarrow{\mathrm{Verifier}}
d_t
\xrightarrow{\mathrm{Update}}
s_{t+1}.
\label{eq:workflow}
\end{equation}
Here \(g_t\) is the proposed subgoal, \(\tau_t\) is the execution trace, and \(d_t\) is the state update decision. The workflow is state-centric: planning reads state to decide what to try, acting attempts the subgoal in the environment, and verification decides what can be written back
into state.

\textbf{Planner.}
The planner reads \(s_t\) and selects the next subgoal \(g_t\). Because the state already exposes current requirements, useful values, and verified evidences, the planner does not need to reconstruct progress from raw history.

\textbf{Actor.}
The actor receives \(g_t\) together with the current state and attempts the subgoal in the environment. It predicts the action trace \(\tau_t\), which contains the next actions taken. Then, the action trace is executed and returns the resulting environment feedback. 

\textbf{Verifier.}
The verifier checks whether the attempted subgoal produced valid progress. It uses structured and agentic probing to examine task-relevant evidence, then outputs the state-update decision \(d_t\). This decision may verify progress, invalidate earlier progress, or keep the relevant state unchanged.

\textbf{Memory.}
We add the memory module as the state-conditioned support. It may suggest useful subgoals, action patterns, or verification checks. Reusable experience helps decide in the above workflow, including the next subgoal and action prediction, and the evidence verification. 


\subsection{Controllable Long-horizon Agentic Planning}
\label{sec:controllable_planning}
The above design naturally makes StructAgent more transparent and controllable, enabling advanced capabilities such as progress checkpointing, failure recovery, and tool-supported execution.


\paragraph{Progress Checkpointing and Resuming.}
The state history naturally serves as a reusable checkpoint for long-horizon execution. Specifically, $s_t^{\mathrm{req}}$ records the minimum sufficient set of current requirements needed to characterize the current task progress, allowing the agent to resume execution without replaying the full interaction history. Meanwhile, $s_t^{\mathrm{val}}$ preserves useful values required by downstream steps, and $s_t^{\mathrm{ver}}$ stores verified evidences supporting verified progress. Together, these components provide a self-contained, human-readable snapshot of the agent's execution state, enabling efficient and reliable workflow resumption.



\paragraph{Failure Recovery and Human Takeover.}
When the state fails to advance, StructAgent recovers from $s_t^{\mathrm{ver}}$ rather than relying on a generic retry loop. The verified evidences identify whether the previous attempt lacked sufficient evidence, conflicted with a verified requirement, repeatedly failed under the current strategy, or was blocked by the environment. Based on this diagnosis, the system selects a targeted intervention, such as replanning, retrying execution, requesting an additional check, repairing the environment, or escalating the task for human takeover when necessary. The selected intervention is inserted as a new subgoal within the same workflow, and any subsequent progress is still committed only through a verifier-backed decision.


\paragraph{Tool-supported execution.}
StructAgent can identify recurring execution patterns from memory and interaction history and summarize them into reusable procedures. In the current runtime, these procedures are exposed through structured memory and typed execution surfaces, each associated with the subgoals it can support, its execution constraints, and the evidence required to verify its outcome. At runtime, the actor can retrieve and invoke these state-conditioned supports to execute $g_t$ more efficiently and reliably, while only verifier-backed decisions can commit their outcomes as valid progress.


\section{Experiments}
\label{sec:exp}

\subsection{Benchmarks and Setup}
\label{sec:setup}

\textbf{Benchmarks.}
Our main evaluation uses OSWorld-Verified
\cite{xlang2025osworldverified}, the verified split of OSWorld
\cite{xie2024osworld} for long-horizon computer use in real desktop
applications. We report the aggregated results in the main table and
provide the full per-domain breakdown in Appendix~\ref{app:osworld_full}.
We also evaluate Mind2Web \cite{deng2023mind2web} as a web-generalization
protocol, grouped by website category. Mind2Web is reported separately because
it uses judge-based trajectory grading rather than executable desktop task
scorers. Finally, we instantiate the same state-centered framework in Minecraft
as a cross-domain generalization study; environment details, task construction,
and additional analysis are provided in Appendix~\ref{app:minecraft}.

\textbf{Compared systems.}
We separate contextual reported results from controlled self-run comparisons.
Reported frontier single-model rows are taken from the OSWorld-Verified
leaderboard \cite{xlang2025osworldverified}. Reported framework rows use the
original results of Agent S3 \cite{agents3_2025}, OS-Symphony
\cite{yang2026ossymphony}, and VLAA-GUI \cite{han2026vlaagui}. These rows
provide reference points under their original settings. Our controlled
comparison is the self-run open-backbone block, where StructAgent and reproduced
single-model, Agent S3, OS-Symphony, and VLAA-GUI baselines are run under
matched model and runner settings.

\textbf{Controlled protocol.}
For OSWorld-Verified, all controlled runs use a 100-step budget and the same
runner and scoring pipeline. We evaluate Qwen3.5-9B and Qwen3.5-27B from the
Qwen vision-language model family \cite{bai2025qwen3vl}, and MiniMax-M3
\cite{lai2026minimaxsparse}. Unless otherwise stated, main results use a single
rollout. Behavior Best-of-N \cite{agents3_2025} is studied separately as a
test-time scaling procedure in Section~\ref{sec:structured_evidence_analysis}.
Prompt schemas in Appendix~\ref{app:prompts}, verification registry details in
Appendix~\ref{app:verification_registry}, and complete runtime configurations
in Appendix~\ref{app:reproducibility} are provided for reproducibility.

\subsection{Main Results on OSWorld-Verified}
\label{sec:osworld}

Table~\ref{tab:main} reports OSWorld-Verified success rates. The table first
lists reported reference results, then compares self-run open-backbone agents
under matched runner settings.

\begin{table}[t]
  \centering\small
  \caption{OSWorld-Verified main results. Success rate (\%) is computed as mean
  Success/Total. The first three rows are reported leaderboard single-model
  results, the next three are reported framework results, and the remaining
  rows are self-run comparisons grouped by open-source backbone. Best values within
  each reported block or self-run backbone group are in \textbf{bold}. Office =
  Calc/Impress/Writer; Daily = Chrome/Thunderbird/VLC; Prof.\ = GIMP/VS\,Code.}
  \label{tab:main}
  \begin{tabularx}{\linewidth}{Y c c c c c c}
    \toprule
    \tablehead System & OS & Office & Daily & Prof. & Multi-Apps & Overall \\
    \midrule
    Claude Sonnet 4.6 & \textbf{91.7} & 75.2 & 76.8 & 70.8 & \textbf{60.2} & 72.1 \\
    Kimi K2.6         & 79.2 & 80.0 & \textbf{77.1} & \textbf{81.6} & 55.0 & 73.1 \\
    Qwen3.7-Plus      & 79.2 & \textbf{82.9} & 74.0 & 73.5 & 59.2 & \textbf{73.3} \\
    
    \midrule
    Agent S3 {\small (Opus/GPT, bBoN)} & 79.2 & 81.1 & 64.9 & 77.6 & \textbf{63.9} & 72.6 \\
    OS-Symphony {\small (GPT-5)}       & 78.3 & 65.7 & 65.9 & 79.6 & 55.3 & 65.8 \\
    VLAA-GUI {\small (Opus 4.5)}       & \textbf{91.7} & \textbf{84.3} & \textbf{72.8} & \textbf{83.7} & 61.1 & \textbf{76.3} \\
    
    \midrule
    \multicolumn{7}{l}{\textbf{Qwen3.5-9B}} \\
    \hspace{1.25em}Single model       & 41.7 & 20.5 & 41.0 & 31.2 & 17.7 & 27.0 \\
    \hspace{1.25em}Agent S3           & 54.2 & 29.9 & \textbf{58.6} & 60.4 & 26.6 & 40.8 \\
    \hspace{1.25em}OS-Symphony        & 56.5 & 33.0 & 55.8 & 58.3 & \textbf{32.6} & 43.3 \\
    \hspace{1.25em}VLAA-GUI           & \textbf{69.6} & 27.8 & 47.9 & \textbf{66.7} & 28.5 & 40.2 \\
    \hspace{1.25em}StructAgent (ours) & 62.5 & \textbf{51.2} & 51.8 & 58.3 & 27.5 & \textbf{46.9} \\
    \cmidrule(lr){1-7}
    \multicolumn{7}{l}{\textbf{Qwen3.5-27B}} \\
    \hspace{1.25em}Single model       & 50.0 & 26.1 & 46.0 & 37.5 & 18.3 & 31.6 \\
    \hspace{1.25em}Agent S3           & 70.8 & 60.6 & 59.5 & 66.7 & \textbf{54.0} & 60.2 \\
    \hspace{1.25em}OS-Symphony        & 70.8 & 47.9 & 58.4 & 64.6 & 43.5 & 52.8 \\
    \hspace{1.25em}VLAA-GUI           & \textbf{83.3} & 30.7 & 56.1 & 62.5 & 47.3 & 48.2 \\
    \hspace{1.25em}StructAgent (ours) & 66.7 & \textbf{71.7} & \textbf{60.1} & \textbf{72.9} & 44.5 & \textbf{62.2} \\
    \cmidrule(lr){1-7}
    \multicolumn{7}{l}{\textbf{MiniMax M3}} \\
    \hspace{1.25em}Single model       & 83.3 & 80.0 & 80.2 & 77.6 & 61.6 & 75.2 \\
    \hspace{1.25em}StructAgent (ours) & \textbf{95.5} & \textbf{84.4} & \textbf{82.9} & \textbf{83.7} & \textbf{69.7} & \textbf{78.9} \\
    
    \bottomrule
  \end{tabularx}
\end{table}

The controlled open-backbone results support three findings.

\textbf{First, StructAgent consistently strengthens the same backbone under
matched runner settings.}
The largest gains appear on weaker open models: Qwen3.5-9B improves from
27.0\% to 46.9\%, and Qwen3.5-27B improves from 31.6\% to 62.2\%. On the
stronger MiniMax-M3 backbone, StructAgent further improves the single-model
baseline and reaches 78.9\%, the strongest open-model result in
Table~\ref{tab:main}. These results suggest that the structured state-centered
loop is not tied to a particular model scale.

\textbf{Second, the gains are largest when task progress can be structured and
verified.}
Office tasks show the clearest improvement, especially on Qwen3.5-9B, where
StructAgent substantially outperforms both the single-model baseline and the
strongest reproduced framework baseline. This is consistent with the design of
StructAgent: spreadsheet and document tasks expose durable intermediate states,
captured values, and application-level checks, allowing the verifier-maintained
state to reduce brittle reliance on raw GUI history. With Qwen3.5-27B, the
advantage becomes broader, with StructAgent leading the 27B group in Office,
Daily, Professional, and Overall, suggesting that stronger backbones can better
use state, evidence, and recovery feedback.

\textbf{Third, cross-application tasks remain the hardest setting.}
StructAgent improves substantially with MiniMax-M3 on Multi-Apps, but the
smaller Qwen3.5-9B backbone still trails OS-Symphony on this group. These tasks
require not only reliable verification, but also reasoning over dependencies
across applications, files, and handoffs. This indicates that state-centered
structure helps preserve progress, while complex cross-application planning
remains a bottleneck for smaller backbones.

\subsection{Web Generalization}
\label{sec:web}

Mind2Web \cite{deng2023mind2web} tests whether the same state-centered design
transfers beyond executable desktop scoring. All self-run systems are evaluated
with the same WebJudge-style protocol \cite{xue2025illusion}, where the judge
receives task points, key screenshots, and action history instead of relying on
self-reported completion.

\begin{table}[t]
  \centering\small
  \caption{Mind2Web generalization results. Success rate (\%) is grouped by
  website category and graded with the same WebJudge-style evaluator. Rows are grouped by open backbone. Ent.\ = Entertainment.}
  \label{tab:mind2web}
  \begin{tabularx}{\linewidth}{Y c c c c c c}
    \toprule
    \tablehead System & Info & Service & Ent. & Shopping & Travel & Overall \\
    \midrule
    \multicolumn{7}{l}{\textbf{Qwen3.5-9B}} \\
    \hspace{1.25em}Agent S3           & 53.8 & \textbf{50.6} & 27.6 & 42.1 & \textbf{35.5} & 46.0 \\
    \hspace{1.25em}OS-Symphony        & 42.9 & 49.4 & 31.0 & 47.4 & 32.3 & 42.6 \\
    \hspace{1.25em}VLAA-GUI           & 44.2 & \textbf{50.6} & \textbf{31.0} & \textbf{52.6} & \textbf{35.5} & 44.3 \\
    \hspace{1.25em}StructAgent (ours) & \textbf{61.5} & 50.0 & 24.1 & 36.4 & 33.3 & \textbf{47.3} \\
    \addlinespace[1pt]
    \multicolumn{7}{l}{\textbf{Qwen3.5-27B}} \\
    \hspace{1.25em}Agent S3           & 61.5 & \textbf{55.3} & 34.5 & 44.7 & \textbf{41.9} & 51.7 \\
    \hspace{1.25em}OS-Symphony        & 49.4 & 51.9 & 37.9 & 52.6 & 38.7 & 47.7 \\
    \hspace{1.25em}VLAA-GUI           & 51.9 & 53.2 & \textbf{44.8} & \textbf{57.9} & \textbf{41.9} & 50.6 \\
    \hspace{1.25em}StructAgent (ours) & \textbf{65.4} & 54.9 & 34.5 & 39.4 & 38.1 & \textbf{52.3} \\
    \bottomrule
  \end{tabularx}
\end{table}

The web results show modest but consistent transfer. With Qwen3.5-9B,
StructAgent reaches 47.3\% overall, slightly above the strongest reproduced
baseline at 46.0\%. With Qwen3.5-27B, StructAgent reaches 52.3\%, again leading
the controlled 27B group. The strongest gains appear in information-seeking
tasks, where progress can often be tracked through searched facts, page states,
URLs, and visible evidence.

The weaker categories clarify the limits of the current implementation.
Entertainment, shopping, and travel tasks often depend on site-specific flows,
dynamic widgets, pop-ups, and transactional choices. These cases create
action-grounding bottlenecks that are not solved by state tracking alone. Thus,
verifier-maintained state helps preserve web progress, but robust browser
exploration and interaction with dynamic sites remain open problems.

\subsection{Minecraft Generalization}
\label{sec:minecraft_exp}

Minecraft is used as a cross-domain instantiation rather than a replacement for
desktop CUA evaluation. The goal is to test whether the same state-centered
structure can be reused when the evidence source changes. In desktop tasks,
verification may rely on screenshots, accessibility trees, files, URLs, or
application-state probes. In Minecraft, the verifier instead uses
inventory-based checks: subgoals are represented as
\((\textsc{verb}, \textsc{item}, n)\), such as
\((\textsc{craft}, \texttt{iron\_pickaxe}, 1)\), and state advances only when
the target inventory condition is verified.The same separation is preserved: the planner, driven by GPT-4o, selects subgoals, specialized actors execute navigation, crafting, or placed-block interactions, and the verifier commits progress only after inventory evidence..

\begin{table}[H]
  \centering\small
  \caption{Minecraft success rate (\%) results. We uses GPT-4o as the backbone model; Optimus-1 numbers are published results from \cite{li2024optimus}.}
  \label{tab:minecraft}
  \begin{tabularx}{\linewidth}{Y c c c c c c}
    \toprule
    \tablehead Agent variant & Wooden & Stone & Iron & Golden & Redstone & Avg. \\
    \midrule
    StructAgent (ours) & \textbf{100} & 70 & \textbf{65} & \textbf{86} & \textbf{57} & \textbf{76} \\
    Optimus-1          & 99 & \textbf{92} & 47 & 9 & 25 & 59 \\
    \bottomrule
  \end{tabularx}
\end{table}

The Minecraft results support the abstraction-level claim: StructAgent's
state-update discipline is not tied to desktop screenshots. It achieves a
76\% task-weighted average success rate, compared with 59\% for Optimus-1,
and improves success on the Iron, Golden, and Redstone tiers, where long
dependency chains make verified subgoal tracking especially useful.
Performance is not uniformly better, however: on Stone tasks, Optimus-1 has a
higher success rate, and StructAgent is sometimes less step-efficient even when
it succeeds. This matches the intended role of the study. It demonstrates that
the same structured state and verifier-committed progress mechanism can be
instantiated with a different evidence source, while leaving domain-specific
actor efficiency as a separate bottleneck.

\section{Analysis}
\label{sec:analysis}

We analyze how the unified state and structured workflow change agent behavior
through four questions: whether verifier-committed state updates prevent false
progress commits, whether structured evidence improves verification, when such
evidence is available, and what failures remain after unified-state progress
control.

\paragraph{Q1: Does the unified state prevent false progress commits?}
\label{sec:state_grounded_analysis}
The unified-state progress control implements StructAgent's core principle: planner
and actor completion claims are proposals, not progress updates. In
long-horizon tasks, a visually plausible screen may still hide a missing file,
an edit to the wrong object, or a later action that overwrites earlier work.
StructAgent therefore advances a subgoal only when a verifier-backed decision
commits it to the unified state.

\begin{figure*}[t]
\centering
\includegraphics[width=\textwidth]{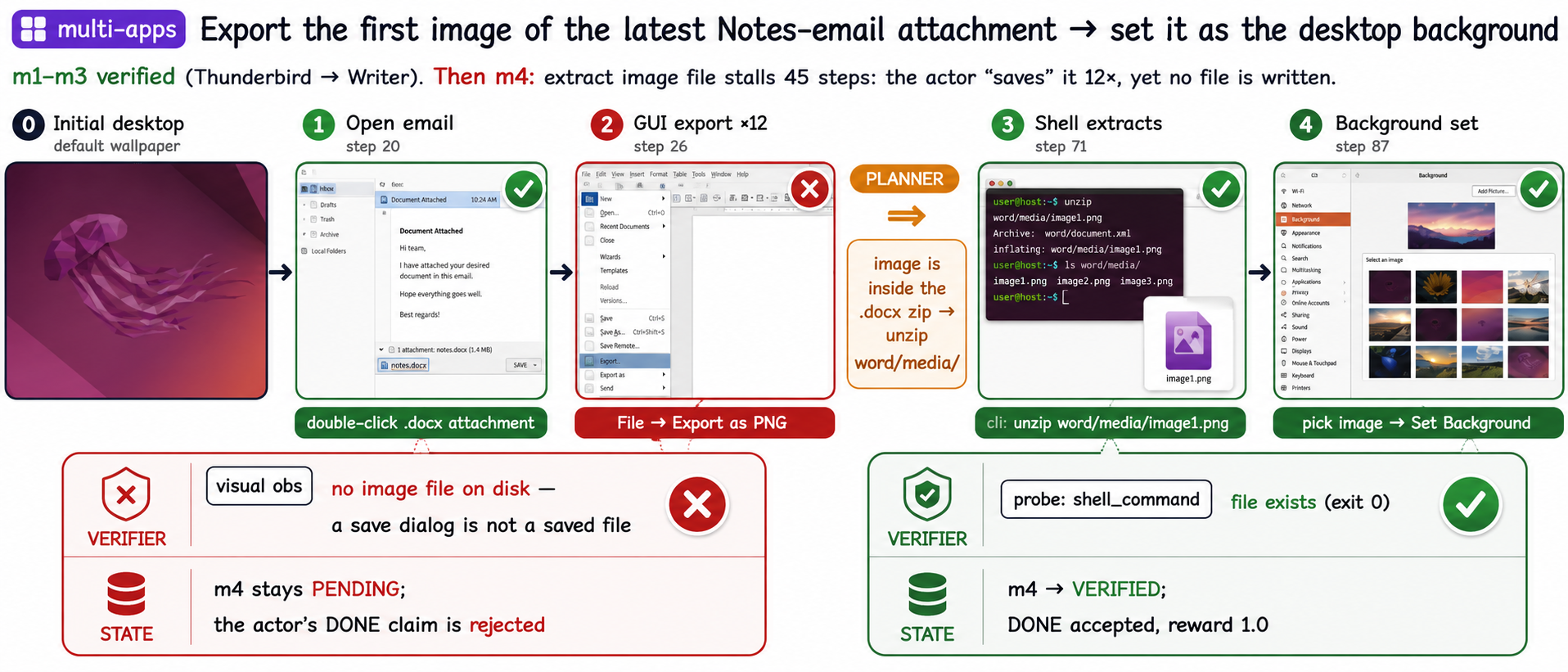}
\caption{Recovery in a multi-app task. The verifier rejects
unsupported GUI-export progress, keeps the subgoal pending, and commits
completion only after shell evidence confirms the extracted image.}
\label{fig:case_study_main}
\end{figure*}

Figure~\ref{fig:case_study_main} illustrates this behavior in a
multi-application task. The actor appears to export an image from a document
attachment, but the verifier finds no matching file evidence and keeps the
image-export subgoal pending. This blocks a false DONE transition and triggers
recovery through a structured probe. Once the extracted image is found on disk,
the verifier commits the subgoal, allowing the planner to safely complete the
wallpaper-setting step.

\paragraph{Q2: Does structured evidence improve verification?}
\label{sec:verification_ablation}
We isolate the verifier on 354 scored Qwen3.5-27B OSWorld-Verified
trajectories, using the OSWorld environment checker as ground truth. We compare
StructAgent's structured verifier with a screenshot-only visual verifier at the
final DONE gate.

\begin{figure}[t]
\centering
\begin{minipage}[t]{0.49\linewidth}
\centering
\includegraphics[width=\linewidth]{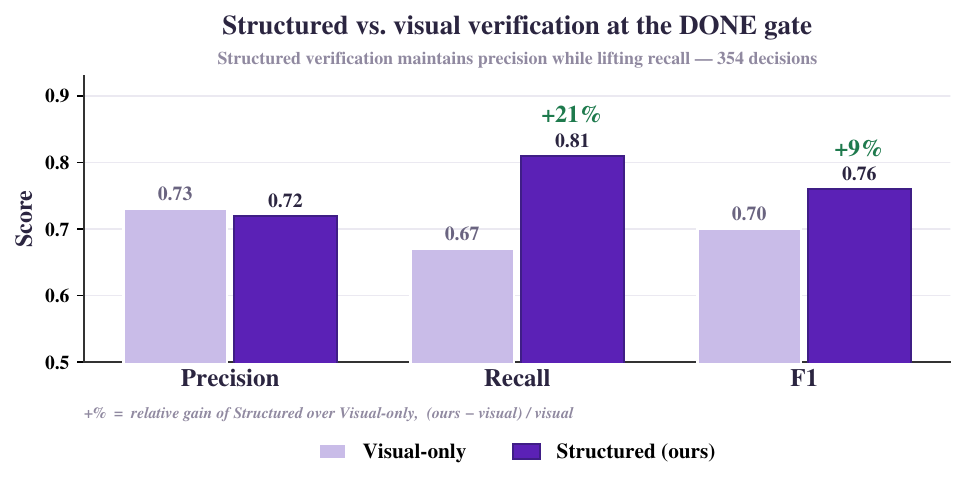}
\vspace{0.2em}
{\footnotesize \textbf{(a)} DONE-gate verification.}
\end{minipage}
\hfill
\begin{minipage}[t]{0.49\linewidth}
\centering
\includegraphics[width=\linewidth]{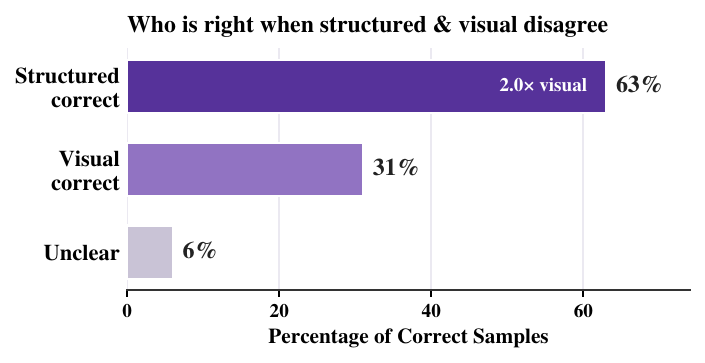}
\vspace{0.2em}
{\footnotesize \textbf{(b)} Disagreement review.}
\end{minipage}
\caption{Structured verification improves final completion checking and is
more often correct when it disagrees with visual-only verification.}
\label{fig:verification_analysis}
\end{figure}

Figure~\ref{fig:verification_analysis}(a) shows that structured verification
raises recall from 0.67 to 0.81 and F1 from 0.70 to 0.76, while keeping
precision close to the visual-only verifier. This indicates that structured
evidence mainly recovers true completions that are difficult to validate from
the final screenshot alone, rather than merely accepting more DONE claims.

We further review 100 intermediate cases where the structured and visual
verifiers disagree, using the instruction, screenshots, verifier evidence, and
probe outputs. As shown in Figure~\ref{fig:verification_analysis}(b),
structured verification is correct in 63\% of disagreements, compared with
31\% for the visual-only verifier and 6\% unclear. Its advantage comes mainly
from latent evidence, such as files, command outputs, URLs, and application
state, that is not reliably visible in screenshots. Visual-correct cases expose
stale probes or proxy matches, suggesting a hybrid verifier that prioritizes
reliable structured checks while retaining visual evidence as a fallback and
cross-check. Representative cases appear in
Appendix~\ref{app:verification_disagreement_cases}.

\paragraph{Q3: When is structured evidence useful?}
\label{sec:structured_evidence_analysis}

\begin{figure}[t]
\centering
\begin{minipage}[t]{0.49\linewidth}
\centering
\includegraphics[width=\linewidth]{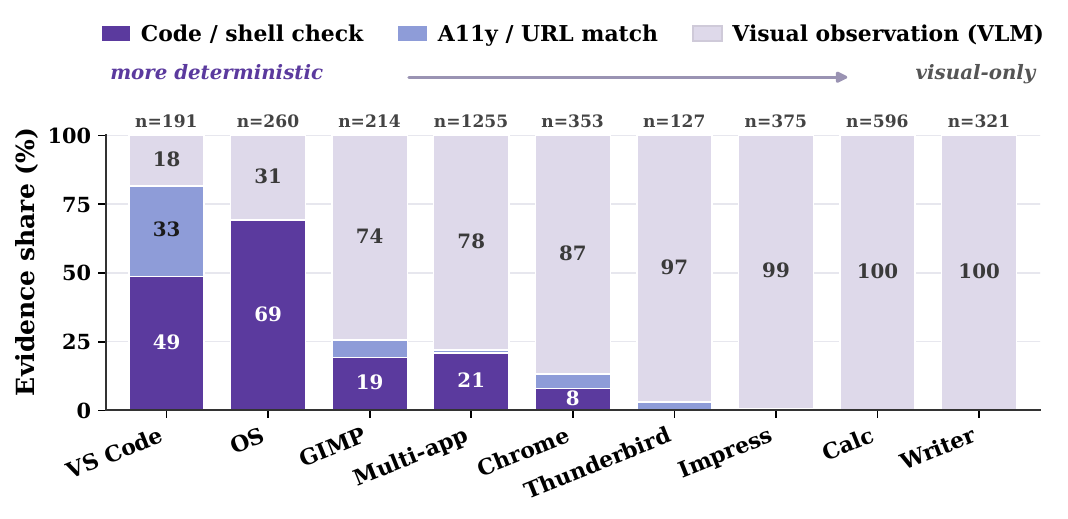}
\vspace{0.2em}
{\footnotesize \textbf{(a)} Runtime evidence sources.}
\end{minipage}
\hfill
\begin{minipage}[t]{0.49\linewidth}
\centering
\includegraphics[width=\linewidth]{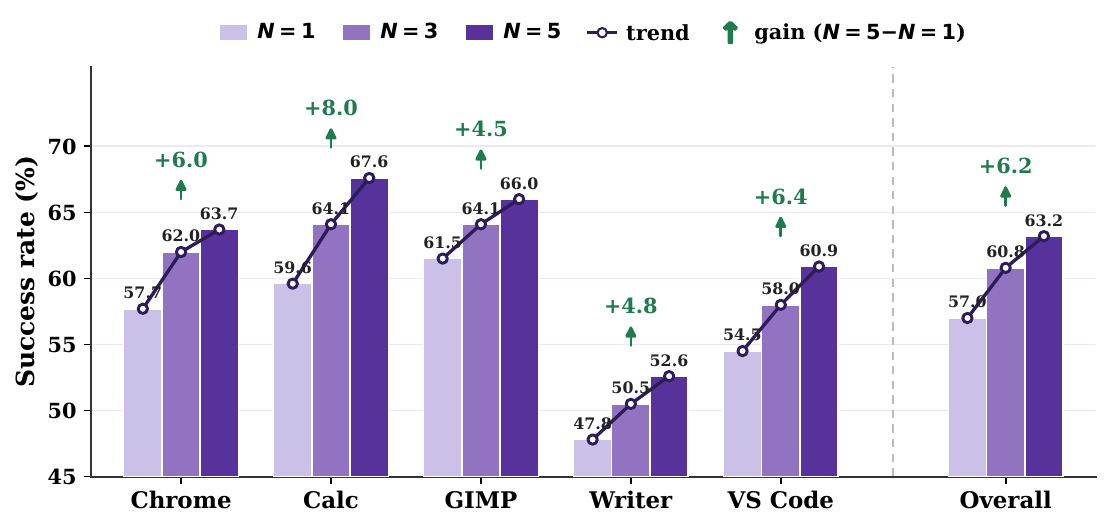}
\vspace{0.2em}
{\footnotesize \textbf{(b)} Test-time BBON scaling.}
\end{minipage}
\caption{Structured evidence at runtime and test time.}
\label{fig:structured_evidence_analysis}
\end{figure}
\vspace{-0.3em}

Figure~\ref{fig:structured_evidence_analysis}(a) shows that structured
verification is most useful when the environment exposes durable task state. VS
Code and OS tasks often provide files, shell outputs, URLs, or accessibility
checks, allowing the verifier to rely less on visual judgment. In contrast,
Calc, Writer, Impress, and several browser or email tasks remain more visual
because success conditions are embedded in GUI state or document layout. Thus,
StructAgent benefits most when task-relevant evidence is externally checkable;
when success is primarily visual, richer app-specific checks and verifier
training become the next bottleneck.

We also test whether this evidence discipline helps at test time. Behavior
Best-of-N (BBON) \cite{agents3_2025} is not a core component of StructAgent, but
we use it to ask whether structured candidate-level evidence helps a verifier
select among completed rollouts. Figure~\ref{fig:structured_evidence_analysis}(b)
shows that moving from one to three rollouts improves every selected domain, and
five rollouts further improves the overall score. Most gains appear early, so
BBON is best viewed as an optional test-time tradeoff.

The domain pattern mirrors runtime verification. Calc, Chrome, and VS Code
benefit more because candidate trajectories expose clearer final state through
cells, pages, files, URLs, or code-visible artifacts. Writer and GIMP improve
less because success is more often tied to visual layout or image-level details.
Structured selection is therefore strongest when candidate evidence aligns with
the benchmark outcome, and weaker when the verifier must fall back to visual
judgment.

\paragraph{Q4: What failures remain?}
\label{sec:failure_analysis}
Finally, we audit 83 score-zero OSWorld-Verified trajectories from the
Qwen3.5-27B run, capped at 20 failures per domain. Each trajectory is assigned a
primary failing role and failure type using saved screenshots, action traces,
verifier logs, and final checker feedback, rather than the agent's own
diagnosis. Failure types include state verification errors, execution failures,
information errors, strategy errors, long-horizon handoff failures, and external
environment blockers.

\begin{figure}[H]
\centering
\includegraphics[width=\linewidth]{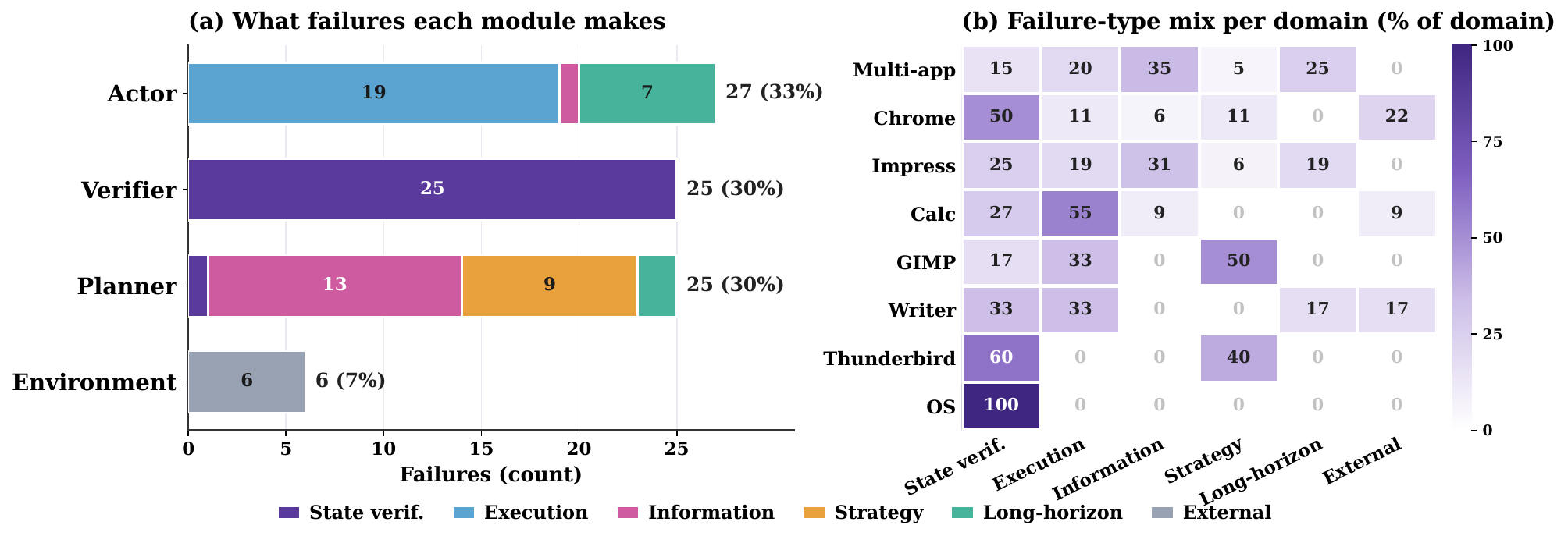}
\caption{Failure analysis on score-zero OSWorld-Verified trajectories. Left:
primary failing role and failure type. Right: domain-level failure-type mix.}
\label{fig:failure_analysis}
\end{figure}

Figure~\ref{fig:failure_analysis} shows that remaining errors are not only actor
grounding failures. Actor failures account for 33\% of audited cases, while
planner and verifier failures each account for 30\%, and environment-side issues
account for 7\%. The largest single failure type is state verification: the
system sometimes accepts weak evidence, misses a completed condition, or fails
to detect that later actions invalidated earlier progress.

The domain breakdown shows that no single module explains all failures. Calc,
GIMP, and Writer are more execution-heavy; Multi-app and Impress expose more
information-selection and handoff failures; Chrome and Thunderbird show more
state-verification or external blockers. This supports the systems view behind
StructAgent: long-horizon reliability requires better evidence design, stronger
verifier training, and domain-aware execution support, not merely scaling one
planner or actor module.
\section{Related Work}
\label{sec:related}

Long-horizon computer-use agents require more than a capable action model: they
need mechanisms for planning over extended tasks, checking whether progress has
actually been made, recovering from partial failures, and reusing reliable
procedures. We therefore organize related work by these structural roles rather
than by individual systems. Table~\ref{tab:related_structure} summarizes the
main design patterns. Prior work often strengthens one part of the agent loop;
StructAgent connects planning, verification, recovery, and tool/skill reuse
through a shared verifier-maintained state.

\begin{table}[h]
  \centering
  \scriptsize
  \setlength{\tabcolsep}{2.4pt}
  \renewcommand{\arraystretch}{1.12}
  \caption{Positioning StructAgent among representative structured-agent
  systems. \textbf{Plan}: long-horizon decomposition or orchestration;
  \textbf{Ver.}: explicit progress or outcome verification;
  \textbf{Rec.}: adaptive recovery or replanning;
  \textbf{T/S}: reusable tools, workflows, or skills;
  \textbf{SS}: shared task-progress state;
  \textbf{VC}: verifier-committed state updates.
  \(\circ\) indicates a partial or auxiliary mechanism.}
  \label{tab:related_structure}
  \begin{tabularx}{\linewidth}{@{}p{0.24\linewidth} Y c c c c c c@{}}
    \toprule
    \tablehead
    \textbf{Line of work} &
    \textbf{Representative works} &
    \textbf{Plan} &
    \textbf{Ver.} &
    \textbf{Rec.} &
    \textbf{T/S} &
    \textbf{SS} &
    \textbf{VC} \\
    \midrule

    Long-horizon agent systems
      & OS-Copilot \cite{wu2024oscopilot}, Cradle \cite{tan2024cradle},
        Agent S2/S3 \cite{agents2_2025,agents3_2025},
        OS-Symphony \cite{yang2026ossymphony}, VLAA-GUI \cite{han2026vlaagui}
      & \textcolor{ForestGreen}{\cmark}
      & \textcolor{gray}{\(\circ\)}
      & \textcolor{ForestGreen}{\cmark}
      & \textcolor{gray}{\(\circ\)}
      & \textcolor{BrickRed}{\xmark}
      & \textcolor{BrickRed}{\xmark} \\

    Progress verification
      & OpenComputer \cite{wei2026opencomputer},
        VAGEN \cite{cui2026vagen},
        GUI-Shepherd \cite{guishepherd2025},
        V-Droid \cite{vdroid2025},
        VeriGUI \cite{zhang2026verigui}
      & \textcolor{BrickRed}{\xmark}
      & \textcolor{ForestGreen}{\cmark}
      & \textcolor{gray}{\(\circ\)}
      & \textcolor{BrickRed}{\xmark}
      & \textcolor{BrickRed}{\xmark}
      & \textcolor{BrickRed}{\xmark} \\

    Memory, tools, and skills
      & Agent Workflow Memory \cite{wang2024awm},
        SkillWeaver \cite{zheng2025skillweaver},
        ToolCUA \cite{hu2026toolcua},
        OS-Copilot \cite{wu2024oscopilot},
        Voyager \cite{wang2023voyager}
      & \textcolor{gray}{\(\circ\)}
      & \textcolor{BrickRed}{\xmark}
      & \textcolor{gray}{\(\circ\)}
      & \textcolor{ForestGreen}{\cmark}
      & \textcolor{BrickRed}{\xmark}
      & \textcolor{BrickRed}{\xmark} \\

    \textbf{StructAgent}
      & \textbf{Unified verifier-maintained state for planning, acting,
        recovery, and tool use}
      & \textcolor{ForestGreen}{\cmark}
      & \textcolor{ForestGreen}{\cmark}
      & \textcolor{ForestGreen}{\cmark}
      & \textcolor{ForestGreen}{\cmark}
      & \textcolor{ForestGreen}{\cmark}
      & \textcolor{ForestGreen}{\cmark} \\

    \bottomrule
  \end{tabularx}
\end{table}

\paragraph{Long-horizon Agent Design.}

Benchmarks for web and desktop control make digital agents measurable on tasks
that require planning, navigation, file editing, cross-application handoffs, and
recovery from partial failures
\cite{xie2024osworld,xlang2025osworldverified,deng2023mind2web,bonatti2024windowsagentarena}.
Recent GUI and computer-use models improve perception and action grounding,
while system-level agents introduce decomposition, specialist roles,
reflection, recovery, tools, or orchestration to make longer tasks more
controllable
\cite{qin2025uitars,xu2025aguvis,wu2024osatlas,wang2025opencua,liu2025scalecua,wu2024oscopilot,tan2024cradle,agents2_2025,agents3_2025,yang2026ossymphony,han2026vlaagui}.
This line motivates our comparison setting: long-horizon reliability is an
agent-design problem, not only a model-scaling problem. StructAgent follows this
systems view, but structures task progress itself as a shared state rather than
only structuring roles, prompts, or execution routes.

\paragraph{Verification and Progress Control.}

A second line studies how to decide whether an agent has actually made progress
or completed a task. Reward and verifier benchmarks show that completion
judgment is fragile when evidence is incomplete, stale, or mostly visual
\cite{lu2025agentrewardbench,lin2025cuarewardbench,sumyk2025arewedoneyet}.
Process-level and environment-state methods address this issue by checking
intermediate actions, hidden application state, or task-specific criteria
instead of trusting self-reported DONE claims
\cite{xiong2025guipra,guishepherd2025,chae2025webshepherd,vdroid2025,wei2026opencomputer,cui2026vagen,rosset2026artverifiers,han2026vlaagui}.
These works are direct comparison cases because StructAgent also treats
unsupported progress as unsafe. The difference is that verification is not only
a reward, critic, STOP gate, or final judge in StructAgent; it is the mechanism
that commits, preserves, or invalidates progress in the shared state.

\paragraph{Memory, Tools, and Skill Reuse.}

Memory and tool-use methods address another long-horizon bottleneck: agents
should not rediscover the same workflows, commands, or interaction patterns from
scratch. Prior work stores reusable experience, induces workflows or skills, and
builds tool surfaces that make execution more reliable across repeated tasks
\cite{wang2024awm,xu2025amem,liu2025cer,sun2025seagent,zheng2025skillweaver,darwinianmemory2026,echotrail2025,sun2026magnet,cheng2025mga,nguyen2025verificagent,zhu2026hymem,hu2026toolcua,wu2024oscopilot}.
Similar ideas also appear in embodied or game environments, where skill
libraries and inventory-level progress provide natural interfaces for long
dependency chains
\cite{wang2023voyager,li2024optimus,lifshitz2023steve1,wang2023jarvis}.
StructAgent also uses memory and tools, but separates reusable advice from
current task truth. A retrieved workflow or tool call can help the actor attempt
a subgoal, while only evidence-backed verifier events can update the state.



\section{Conclusion}
\label{sec:conclusion}
We presented \textbf{StructAgent}, a state-centered framework for long-horizon computer use based on two complementary designs: a unified state that provides a compact and verifiable representation of task progress, and a structured workflow that regulates progress through verifier-backed state transitions. Rather than treating long-horizon execution as continuous generation over an ever-growing interaction history, StructAgent organizes planning, acting, and verification around an explicit causal progress structure. This design enables transparent progress checkpointing, evidence-driven task completion, targeted failure recovery, and tool-assisted execution while preserving a consistent and auditable working process.
Extensive experiments demonstrate that StructAgent consistently improves diverse LLM and VLM backbones on long-horizon computer-use benchmarks, achieving a new open-source state-of-the-art on OSWorld-Verified and generalizing beyond desktop environments to Minecraft.

Looking forward, we believe the proposed state-centered paradigm can naturally extend beyond computer use to broader agentic systems, including web agents, embodied agents, and multi-agent collaboration. Future work may further enrich the state with richer semantic and causal representations, learn adaptive workflow policies, and investigate scalable verification mechanisms for increasingly complex real-world tasks.
\printbibliography

\clearpage

\appendix

\section{Additional OSWorld Results and Ablations}
\label{app:osworld_full}

Table~\ref{tab:osworld_domain_detail} reports the full 10-domain OSWorld
breakdown for controlled self-run rows. Scores are computed from the summed
task scores divided by the number of scored tasks in each domain, so partial
credit is preserved when the benchmark scorer returns it.

\begin{table}[h]
  \centering\scriptsize
  \caption{OSWorld-Verified per-domain self-run results (success rate \%).
  Chr. = Chrome; Th. = Thunderbird; VS = VS Code.}
  \label{tab:osworld_domain_detail}
  \resizebox{\linewidth}{!}{%
  \begin{tabular}{llccccccccccc}
    \toprule
    \tablehead System & Backbone & Chr. & GIMP & Calc & Impress & Writer &
    Multi & OS & Th. & VLC & VS & Overall \\
    \midrule
    StructAgent (ours) & Qwen3.5-9B & 57.7 & 61.5 & 59.6 & 44.5 & 47.8 & 27.5 & 62.5 & 60.0 & 28.9 & 54.5 & 46.9 \\
    Single model & Qwen3.5-9B & 41.2 & 26.9 & 8.5 & 27.5 & 30.4 & 17.7 & 41.7 & 60.0 & 23.5 & 36.4 & 27.0 \\
    Agent S3 & Qwen3.5-9B & 60.8 & 50.0 & 29.8 & 27.5 & 34.8 & 26.6 & 54.2 & 71.4 & 41.3 & 72.7 & 40.8 \\
    OS-Symphony & Qwen3.5-9B & 52.1 & 65.4 & 32.6 & 25.4 & 50.0 & 32.6 & 56.5 & 71.4 & 52.9 & 50.0 & 43.3 \\
    VLAA-GUI & Qwen3.5-9B & 54.3 & 69.2 & 21.3 & 20.1 & 56.5 & 28.5 & 69.6 & 46.7 & 30.7 & 63.6 & 40.2 \\
    \addlinespace[1pt]
    StructAgent (ours) & Qwen3.5-27B & 60.8 & 76.9 & 76.6 & 65.8 & 73.9 & 44.5 & 66.7 & 66.7 & 51.6 & 68.2 & 62.2 \\
    Single model & Qwen3.5-27B & 47.7 & 30.8 & 15.2 & 32.6 & 34.8 & 18.3 & 50.0 & 60.0 & 28.9 & 45.5 & 31.6 \\
    Agent S3 & Qwen3.5-27B & 54.3 & 61.5 & 72.3 & 42.3 & 73.9 & 54.0 & 70.8 & 66.7 & 67.5 & 72.7 & 60.2 \\
    OS-Symphony & Qwen3.5-27B & 52.1 & 53.8 & 48.9 & 36.2 & 69.6 & 43.5 & 70.8 & 80.0 & 56.2 & 77.3 & 52.8 \\
    VLAA-GUI & Qwen3.5-27B & 45.6 & 46.2 & 31.9 & 16.8 & 56.5 & 47.3 & 83.3 & 80.0 & 63.5 & 81.8 & 48.2 \\
    \addlinespace[1pt]
    Single model & MiniMax-M3 & 76.0 & 76.9 & 80.9 & 73.7 & 91.2 & 61.6 & 83.3 & 86.7 & 86.6 & 78.3 & 75.2 \\
    StructAgent (ours) & MiniMax-M3 & 73.8 & 84.6 & 89.1 & 74.3 & 95.6 & 69.7 & 95.5 & 93.3 & 98.2 & 82.6 & 78.9 \\
    \bottomrule
  \end{tabular}}
\end{table}

\section{Web Benchmark Details}
\label{app:web_details}

\paragraph{Benchmarks.}
We prepare a Mind2Web \cite{deng2023mind2web} web-generalization evaluation.
Results are grouped by website category: information, service, entertainment,
shopping, and travel.

\paragraph{Mind2Web evaluation.}
For Mind2Web, we follow the evaluation protocol and prompt design of WebJudge
\cite{xue2025illusion}. The judge receives key task points, key screenshots, and
the action history, and decides whether the trajectory satisfies the user
instruction. We use Qwen3.5-27B \cite{bai2025qwen3vl} as the judge model for the
Mind2Web rows. This keeps the comparison under a shared evaluation protocol
rather than relying on framework-specific success declarations.

\begin{table}[h]
  \centering\small
  \caption{Reliability of the Mind2Web judge used in our evaluation. Agreement is
  measured against human labels on a held-out validation set.}
  \label{tab:mind2web_judge_reliability}
  \begin{tabularx}{\linewidth}{Y c c c c c}
    \toprule
    \tablehead Judge & $N$ & Agreement (\%) & SR Gap (pp) & False Pos. (\%) & False Neg. (\%) \\
    \midrule
    Qwen3.5-27B WebJudge-style evaluator & 150 & 86.0 & 4.7 & 4.7 & 9.3 \\
    \bottomrule
  \end{tabularx}
\end{table}

On this validation set, the judge agrees with human labels on 129 of 150
examples, with 7 false positives and 14 false negatives. This corresponds to a
4.7 point success-rate gap against human labels. The lower false-positive rate
indicates a conservative evaluator, which is preferable for Mind2Web because
over-crediting failed trajectories would inflate the reported success rate.

\paragraph{Reported and self-run rows.}
Self-run rows use the same evaluation code and judge model within each benchmark.
Published reference rows, when included, follow the original benchmark settings
and are shown only as external reference points.

\section{State, Verification Registry, and Execution Surfaces}
\label{app:verification_registry}

\paragraph{State schema.}
The implementation stores task progress in a typed \texttt{VerificationRecord}.
The main milestone object is \texttt{Outcome}. Each outcome contains an id,
description, evidence hint, dependencies, an optional app tag for multi-app
tasks, an optional \texttt{VerifySpec}, and an authoritative state from
\texttt{pending}, \texttt{verified}, and \texttt{reverted}, where
\texttt{reverted} is the implementation counterpart of \textsc{Invalidated}. Planner and actor
outputs are recorded in the timeline, but they do not directly change an
outcome state.

\begin{table}[t]
  \centering\scriptsize
  \caption{Implementation-level state components.}
  \label{tab:state_schema_app}
  \begin{tabularx}{\linewidth}{p{0.22\linewidth} Y p{0.18\linewidth} Y}
    \toprule
    \tablehead Object & Main fields & Updated by & Purpose \\
    \midrule
    \texttt{VerificationRecord} & \texttt{initial\_context},
    \texttt{required\_outcomes}, legacy \texttt{done}, \texttt{failed\_paths},
    \texttt{completed\_strategies}, \texttt{slots}, and
    \texttt{global\_facts}. & Initializer, verifier, planner loop & Provides
    the shared task state rendered back to the planner. \\
    \texttt{Outcome} & \texttt{id}, \texttt{description},
    \texttt{evidence\_hint}, \texttt{depends\_on}, \texttt{verify},
    \texttt{app}, \texttt{state}, rejection counters, facts, evidence, and
    failures. & Initializer and verifier & Represents one durable gating
    milestone. \\
    \texttt{VerifySpec} & One structured verifier specification attached to an
    outcome, such as \texttt{file\_grep} or \texttt{calc\_verify}. &
    Initializer and verifier rewrite logic & Defines how an outcome can be
    checked without free-form code generation. \\
    \texttt{StepRecord} and \texttt{TimelineEvent} & Screenshot, a11y text,
    URL, actor response, emitted action code, planner decision, key nodes, and
    perceiver snapshot. & Runtime loop & Preserves the execution history used
    by planner rendering, verifier context, debugging, and audit. \\
    \texttt{Evidence}, \texttt{Fact}, and \texttt{Failure} & Verifier traces,
    bound slot values, extracted facts, and failed strategy or action records.
    & Verifier, extraction handlers, failure attribution & Carries grounded
    useful values and recovery evidences across later turns. \\
    \bottomrule
  \end{tabularx}
\end{table}

\paragraph{Actor-loop and boundary verification.}
In the current boundary-verification mode, per-step KeyNode polling is bypassed.
The runtime still records each \texttt{StepRecord}, but milestone verification
runs when the actor loop returns control to the planner, and once more when the
planner claims \texttt{DONE}. The boundary verifier receives the current
screenshot, up to four prior screenshots from the timeline, prefiltered a11y
text, the last eight actor responses, the current subgoal, the target outcomes,
and optional verifier-memory check recipes. If a target is verified, the
corresponding \texttt{Outcome} is marked \texttt{verified} and stores a
\texttt{last\_verifier\_trace} with source, method, reason, evidence,
confidence, step index, and the authored check when one was used. Unverified or
uncertain outcomes remain in their previous state and are reported back to the
planner.

The low-level actor verifier is separate from milestone verification. It does
not update the milestone state or verified evidences. It compares the before and after screenshots for the
current actor turn, with up to three prior actor-turn screenshots and the last
three rendered action-history lines. Its statuses are \texttt{complete},
\texttt{continue}, \texttt{no\_effect}, \texttt{off\_track}, and
\texttt{stuck}; these statuses decide whether actor control continues or returns
to the planner.

\paragraph{Verification registry.}
The boundary verifier can either judge from observation or request one
validated \texttt{VerifySpec}. The actual probe kinds are the seven kinds below.
They return \texttt{True}, \texttt{False}, or \texttt{None}; a determinate
result overrides the observation judgment, while \texttt{None} falls back to
observation.

\begin{table}[t]
  \centering\scriptsize
  \caption{Actual \texttt{VerifySpec} kinds used by the verification registry.}
  \label{tab:verification_registry_app}
  \begin{tabularx}{\linewidth}{p{0.18\linewidth} Y Y}
    \toprule
    \tablehead Kind & Required fields and evidence & Typical use \\
    \midrule
    \texttt{file\_grep} & \texttt{file\_path}, regex \texttt{patterns},
    optional \texttt{all\_must\_match} and \texttt{proximity\_lines}; reads a
    VM file and matches text. & Plaintext files, saved configs, bookmarks, and
    exported text artifacts. \\
    \texttt{url\_match} & \texttt{url\_pattern}; extracts the active browser URL
    from address-bar a11y rows. & Chrome navigation and page-open conditions. \\
    \texttt{a11y\_match} & Optional \texttt{tag}, \texttt{text\_contains},
    \texttt{state\_contains}, and visual clauses. & Visible UI state, controls,
    dialogs, forms, Thunderbird, VLC, and other live app states. \\
    \texttt{calc\_verify} & A list of \texttt{calc\_checks} whose ops are
    validated against the Calc action library. & Spreadsheet cells, formulas,
    formats, sorting, frozen panes, and related Calc state. \\
    \texttt{writer\_verify} & A list of \texttt{writer\_checks} validated by the
    Writer action library. & Document text, paragraph properties, and character
    formatting. \\
    \texttt{impress\_verify} & A list of \texttt{impress\_checks} validated by
    the Impress action library. & Slide text, shapes, notes, and presentation
    formatting. \\
    \texttt{shell\_command} & An argv list plus expected or forbidden
    substrings. Boundary probes are guarded by a read-only allowlist such as
    \texttt{which}, \texttt{pgrep}, \texttt{test}, \texttt{gsettings get},
    \texttt{dpkg -l}, and \texttt{snap list}. & File existence, process state,
    installed packages, and system settings. \\
    \bottomrule
  \end{tabularx}
\end{table}

\paragraph{Registry selection and fallback.}
The prompt exposes only the probe kinds allowed by the current domain's trust
map. For example, Chrome exposes \texttt{a11y\_match}, \texttt{url\_match}, and
\texttt{file\_grep}; Calc exposes \texttt{calc\_verify} and
\texttt{file\_grep}; Writer and Impress expose their corresponding UNO-backed
checks plus \texttt{file\_grep}; OS tasks expose \texttt{file\_grep} and
\texttt{shell\_command}. The verifier may return a \texttt{probe} object, but
\texttt{spec\_from\_probe} accepts it only if the kind is allowed, the fields
are whitelisted, the spec is valid, and any shell command passes the read-only
guard. Invalid probes are ignored. If a valid probe returns \texttt{True} or
\texttt{False}, the method is recorded as \texttt{probe:<kind>}; if it returns
\texttt{None}, the method is recorded as \texttt{probe-inconclusive->obs}.

\paragraph{Execution surfaces.}
The actor emits actions from the decomposer allowlist. These actions may change
the environment, but they do not mark progress unless boundary verification
later updates an outcome.

\begin{table}[t]
  \centering\scriptsize
  \caption{Execution surfaces and their verification paths.}
  \label{tab:execution_surfaces_app}
  \begin{tabularx}{\linewidth}{p{0.20\linewidth} Y Y}
    \toprule
    \tablehead Surface & Actual actor actions & Verification path \\
    \midrule
    GUI / pyautogui & \texttt{click}, \texttt{left\_double},
    \texttt{right\_single}, \texttt{drag}, \texttt{type}, \texttt{hotkey},
    \texttt{scroll}, and \texttt{wait}. Coordinates are emitted inline on a
    0--1000 grid and converted to pixels. & Low-level VLM verifier,
    screenshots, a11y, and later boundary probes. \\
    Browser direct action & \texttt{navigate(url, new\_tab)} through the Chrome
    DevTools interface, plus \texttt{finished(answer)} as a text-answer terminal
    sentinel. Page interaction still uses GUI actions. & \texttt{url\_match},
    \texttt{a11y\_match}, screenshots, and task-specific final judging. \\
    Host or VM handlers & \texttt{cli\_run}, \texttt{edit\_json},
    \texttt{extract\_info}, and \texttt{open\_app}. These
    are actor-side tools and can change task state when used by the actor. &
    Boundary verifier checks the resulting file, app, browser, or verification-record state;
    verifier-side \texttt{shell\_command} remains a separate read-only probe. \\
    Structured Office actions & The allowlist includes validated
    \texttt{calc\_*}, \texttt{writer\_*}, and \texttt{impress\_*} actions,
    lowered by their action libraries to runtime scripts. &
    \texttt{calc\_verify}, \texttt{writer\_verify},
    \texttt{impress\_verify}, file checks, and visual fallback. \\
    Terminal actor claims & \texttt{impossible} maps to an
    \texttt{IMPOSSIBLE} sentinel; planner-level \texttt{DONE} is accepted only
    after the verification gate and DONE auditor. & Feasibility logic, boundary
    verification on pending leaves, and the DONE auditor. \\
    \bottomrule
  \end{tabularx}
\end{table}

\paragraph{Safety, logging, and reproducibility.}
Verifier-side shell probes are read-only guarded and cannot use shell pipelines,
redirects, or free-form interpreters. Actor-side commands such as
\texttt{cli\_run} are separate execution actions and are logged through the
normal action history. Boundary-verifier debug files record allowed probes,
target outcomes, verdict methods, reasons, evidence, and authored checks; the
timeline retains the screenshots, a11y text, actor responses, and action code
needed to audit those decisions. Full prompts and structured output schemas are
provided in Appendix~\ref{app:prompts}.

\section{Structured Memory and Tool Construction}
\label{app:memory_tools}

\paragraph{Memory as reusable advice, not task truth.}
StructAgent separates memory from state. Memory stores reusable experience from
past trajectories: plans that worked, common action patterns, app-specific
rules, pitfalls, and verifier check recipes. The state stores what is currently
true in the running task. A memory entry can suggest what to try or how to
check a milestone, but it cannot mark an outcome as complete. Outcome state is
updated only by verifier-backed decisions, as described in
Appendix~\ref{app:verification_registry}.

\paragraph{Offline construction from trajectories.}
The memory bank is built offline from three trajectory sources. The internal
source is produced by running our agent on CUA-Gym \cite{wang2026cuagym} tasks.
These rollouts are launched by
\texttt{scripts/run\_rollout\_subbatch.sh}, adapted into the same execution
runner, and mined into
\texttt{results/planner\_experience/<domain>/<task\_id>.json}. The external
sources are normalized AgentNet desktop trajectories from OpenCUA
\cite{wang2025opencua}, stored under \texttt{results/agentnet\_normalized},
and normalized Multimodal-Mind2Web \cite{deng2023mind2web} web trajectories,
stored under
\texttt{results/mind2web\_normalized}. The loader converts all three sources
into the same \texttt{UnifiedRecord} schema with instruction, domain, subgoals,
key actions, outcomes, and pitfalls. The unified pool is clustered at task and
subgoal levels and then polished into several reusable surfaces. Task-level
clusters become plan templates; subgoal segments become typical-action
memories; loose clusters become general rules and pitfalls; domain summaries
become app-level rule sheets. A leakage-cleaning pass abstracts task-instance
values such as file names, URLs, product names, and exact answers while
preserving reusable procedures. The resulting entries are indexed with
\texttt{sentence-transformers/all-MiniLM-L6-v2}, based on Sentence-BERT
\cite{reimers2019sentencebert} and MiniLM \cite{wang2020minilm}, and FAISS
\cite{johnson2019billion}. Verifier-side check recipes are built separately
from successful verifier intent recipes and indexed by their
\texttt{when\_to\_use} field.

\begin{table}[t]
  \centering\scriptsize
  \caption{Structured memory surfaces and where they enter the runtime loop.}
  \label{tab:memory_surfaces_app}
  \begin{tabularx}{\linewidth}{p{0.19\linewidth} p{0.22\linewidth} p{0.22\linewidth} Y}
    \toprule
    \tablehead Runtime point & Memory source & Retrieval key & Runtime use \\
    \midrule
    Task start & L2 plan templates & Task instruction through
    \texttt{when\_to\_use}. & Similar full plans, typical subgoals, and
    task-level pitfalls for the initial planner. \\
    Task start & L3a cluster summaries & Task instruction through
    \texttt{cluster\_summary}. & Broad task-class patterns shared across
    related tasks. \\
    Task start & L3c domain rules & Direct domain lookup. & Stable app facts,
    common widgets, shortcuts, and domain-level pitfalls. \\
    Subgoal transition & L1 typical actions & Current subgoal and domain
    through \texttt{subgoal\_pattern}. & Concrete action recipes and
    tools/widgets seen for similar subgoals. \\
    Subgoal transition & L3a individual rules & Current subgoal through
    \texttt{rule} and \texttt{applies\_when}. & Local gotchas and reusable
    constraints for the next action. \\
    Boundary verification & Check recipes & Subgoal, target milestone text,
    and domain through \texttt{when\_to\_use}. & Guidance for probe choice,
    strict pass criteria, and known verifier traps. \\
    \bottomrule
  \end{tabularx}
\end{table}

\paragraph{Task-level planning memory.}
At task start, the planner retrieves L2, L3a, and L3c memory. L2 plan
templates provide concrete subgoal sequences from similar successful tasks. L3a
cluster summaries provide broader patterns and shared pitfalls for the task
class. L3c domain rules are loaded by domain rather than by semantic search;
they capture stable facts about an application, such as common controls,
shortcuts, and failure modes. These entries are rendered as planning advice, not
as required outcomes.

\paragraph{Subgoal-level action memory.}
During subgoal transitions, the actor-side prompt receives L1 typical actions
and L3a individual rules. L1 entries are mined from subgoal segments and contain
typical actions, approaches, tools or widgets seen, and common pitfalls. L3a
rules give broader guidance that still applies locally, such as avoiding a
known brittle path or preferring a reliable app-native control. These memories
help the actor choose a concrete execution path, but the effect still has to be
checked by low-level and boundary verification.

\paragraph{Verifier-side check recipes.}
Verifier memory is separate from planner and actor memory. The offline builder
distills old verifier intent recipes into boundary check recipes that describe
which probe kind is reliable for a milestone class, what a strict pass condition
should require, and which signals are misleading. At runtime,
\texttt{retrieve\_check\_recipes} returns the top recipes for the current
subgoal and target milestones. The retrieved recipes are guidance only: the
boundary verifier must still author a valid \texttt{VerifySpec}, and that spec
must pass the registry validation and read-only guards before it can run.

\paragraph{Trajectory-derived tools and typed execution surfaces.}
StructAgent follows the tool-CUA idea of extracting reusable procedures from
past trajectories \cite{hu2026toolcua}, but the current runtime does not rely
on online synthesis of new executable tools. Reusable procedures appear as
structured memory entries, and execution happens through typed action surfaces.
The active actor-side surfaces include GUI actions, \texttt{navigate},
\texttt{cli\_run}, \texttt{edit\_json}, \texttt{extract\_info},
\texttt{open\_app}, and validated \texttt{calc\_*}, \texttt{writer\_*}, and
\texttt{impress\_*} actions. These surfaces reduce brittle low-level GUI
operation, but they still only produce actions; verifier events decide whether
the task state has advanced.

\paragraph{Safety and logging.}
Memory retrieval is optional and failure-safe: missing indexes or retrieval
errors yield an empty memory block rather than changing the control flow. The
runtime dumps rendered memory blocks and retrieval metadata under the results
directory for post-hoc inspection. Leakage cleaning and structured rendering
keep memory focused on reusable procedures instead of task-specific answers,
and no memory entry can directly accept a DONE claim.

\begin{figure}[h]
  \centering
  \includegraphics[width=\textwidth]{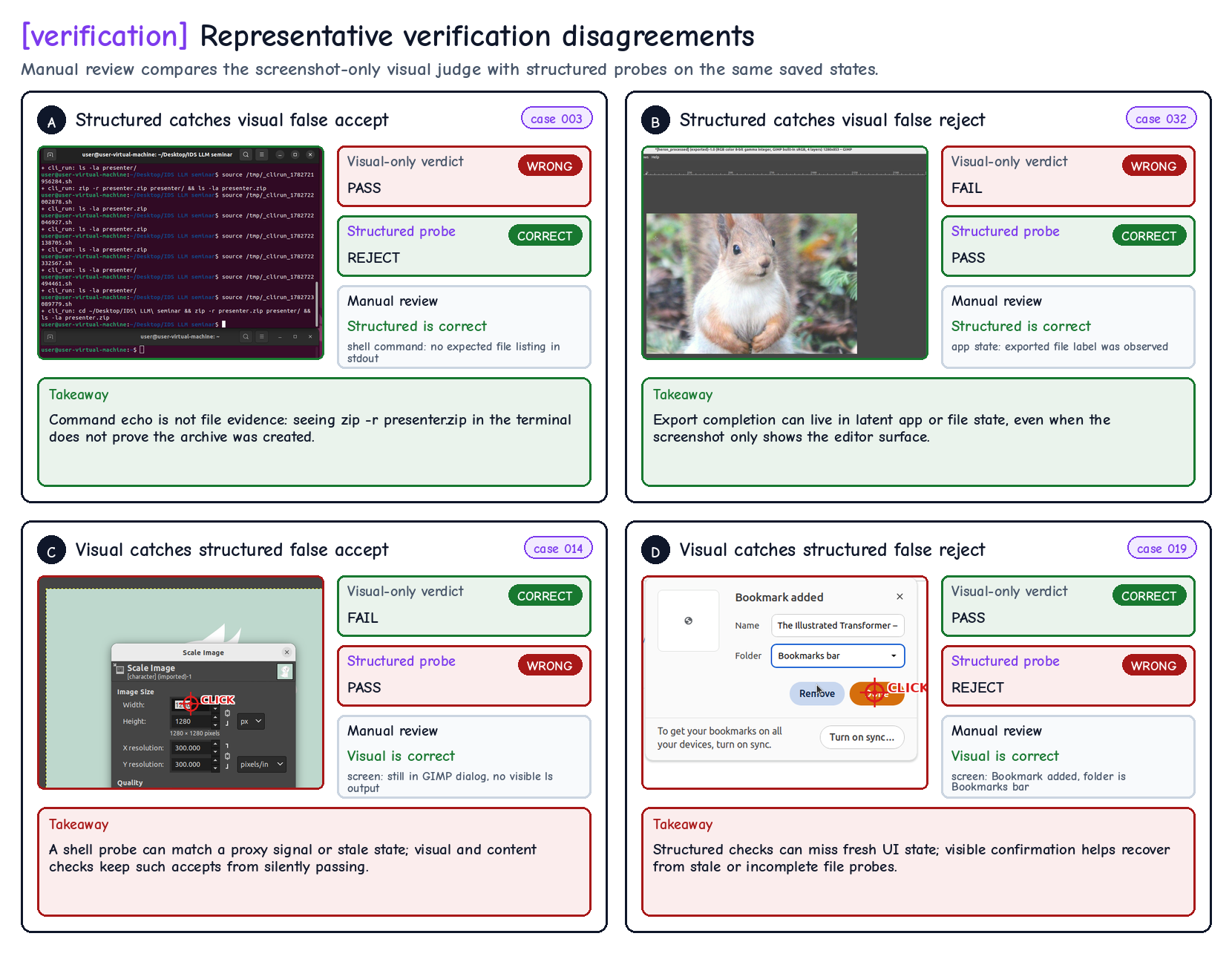}
  \caption{Representative verification disagreements. Structured probes recover
  latent file or app evidence missed by screenshots, while visual evidence
  catches stale or proxy probe matches.}
  \label{fig:verification_disagreement_cases}
\end{figure}
\section{Additional Analysis and Case Studies}
\label{app:additional_analysis}

\subsection{DONE Auditing Prevents Premature Completion}
\label{app:done_auditor_case_study}

\begin{figure}[h]
  \centering
  \includegraphics[width=\textwidth]{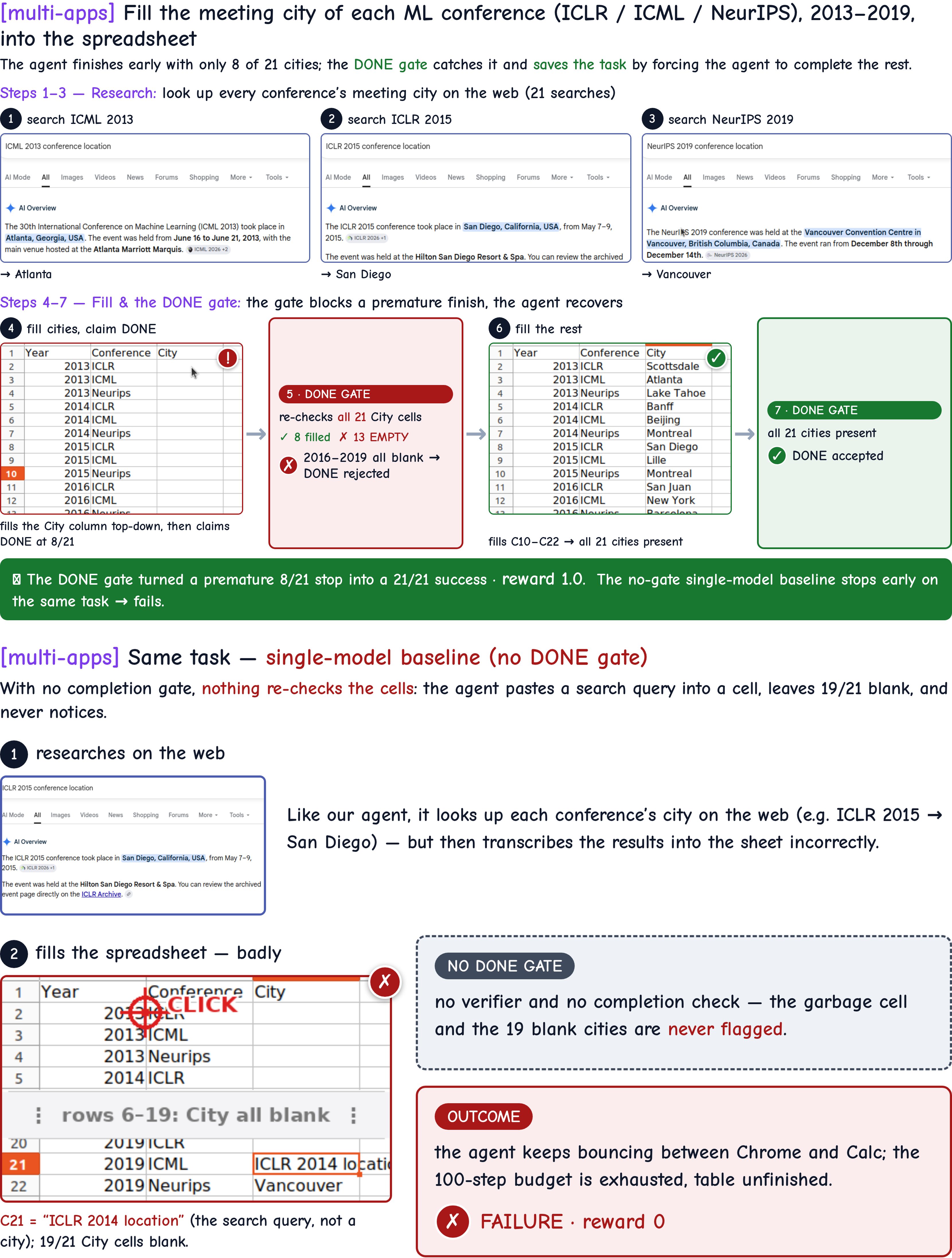}
  \caption{DONE auditing in a multi-app spreadsheet task. Top: the auditor
  rejects an early 8/21 completion claim and accepts the task only after all 21
  city cells are filled. Bottom: without a DONE gate, the baseline never flags
  blank cells or a copied search query and exhausts the step budget.}
  \label{fig:case_study_done_auditor}
\end{figure}

Figure~\ref{fig:case_study_done_auditor} shows a multi-application spreadsheet
task where the agent must fill the meeting city for each ICLR, ICML, and NeurIPS
conference from 2013 to 2019. In StructAgent, a DONE claim is only a proposal.
The DONE auditor
rechecks the required city cells, rejects an early 8/21 completion claim, keeps
the task open, and accepts the task only after all 21 cities are present. This
case illustrates why completion auditing is useful for long table-filling tasks:
local actions can look reasonable while the global task state is still
incomplete. Without this gate, the baseline performs some correct web lookups,
but no module checks the final spreadsheet as a whole: one cell contains a
copied search query, many city cells remain blank, and the run exhausts its step
budget.

\subsection{Representative Verification Disagreements}
\label{app:verification_disagreement_cases}

Figure~\ref{fig:verification_disagreement_cases} shows four manually reviewed
cases where structured and visual verification disagree. Structured probes catch
visual false accepts and false rejects when the relevant evidence lives in
terminal output, file state, or application state. Visual evidence remains useful
when a probe matches a proxy signal or misses a fresh UI confirmation.

\subsection{Additional BBON Details}
\label{app:bbon}
We use Behavior Best-of-N \cite{agents3_2025} as a test-time, verifier-side
analysis, not as part of the agent core. For each task we take $N{=}5$ completed
rollouts and render each as a behavior narrative: turning-point screenshots,
the per-step accessibility state
(URL, active tab, element states) paired with each frame, and the final structured
probe (for spreadsheets: cell values, formulas, and charts; for Chrome: page and
DOM state). A single judge (Qwen3.5-27B \cite{bai2025qwen3vl}, structured
per-candidate prompt, temperature 0) then selects the most credible trajectory.
The judge never sees the ground-truth score; we evaluate its pick afterwards. The main finding is that
judge quality is bounded by \emph{evidence}, not reasoning: feeding checker-aligned
evidence, such as structured final state and per-step accessibility text, is what
turns BBON from harmful or near-random into a positive gain.

\section{Minecraft Generalization}
\label{app:minecraft}

\subsection{Environment}
\label{app:mc_env}
We use the MineRL simulator \cite{guss2019minerl} with Minecraft 1.16.5, the
same setting as Optimus-1 \cite{li2024optimus}. The agent receives a $128 \times 128$ RGB observation
and emits a 23-element discrete-plus-continuous action dictionary at a fixed
20\,Hz tick rate. Episodes terminate on milestone satisfaction or on a
configurable step budget (10--15 minutes wall time, equivalent to
12{,}000--18{,}000 environment steps). All planning and State-decomposition calls use GPT-4o, while the actors (STEVE-1 and the JARVIS-1 craft/smelt helpers) are separate specialized models, so the reasoning model is held fixed and only the evidence source changes across domains.

\subsection{Task suite}
\label{app:mc_tasks}
Following the Optimus-1 task setting \cite{li2024optimus}, tasks are grouped
into seven tiers ordered by canonical crafting depth:
Wooden (12 tasks), Stone (10), Iron (17), Golden (7), Redstone (7),
Diamond (7), Armor (13). Tier depth corresponds to plan length: a Wooden
task is reachable in $\le 5$ sub-tasks under empty inventory; an Iron task
requires $\approx 11$ sub-tasks; Golden and Redstone reach $\approx 13$
sub-tasks. We evaluated Wooden through Redstone.

\subsection{Memory bank}
\label{app:mc_memory_bank}
The Minecraft memory bank is a recipe-grounded knowledge graph that
supplies the planner with canonical sub-task chains for any target item.
It is built once offline from three sources: (i) the Minecraft 1.16.5
client recipe data, giving authoritative ingredient slots and counts;
(ii) the JARVIS-1 \parencite{wang2023jarvis} sub-task ontology for
non-recipe actions (\texttt{mine}, \texttt{smelt}, \texttt{kill});
(iii) verified successful trajectories from the Optimus-1 experience
pool \parencite{li2024optimus}, used to align canonical chains with the
benchmark.

The bank is a hierarchical directed acyclic graph with 613 item nodes
and 848 typed edges; nodes are items, edges encode \texttt{produces}
with ingredient multiplicity. Each node carries a tier attribute
$\in \{\textsc{wood, stone, iron, gold, redstone, diamond, armor}\}$
from the shortest dependency path. At construction time, every
craftable item is compiled to a canonical plan via topologically
ordered DFS to mining-source leaves, materialized as a list of
$(\textsc{verb}, \textsc{item}, n)$ tuples matching the state
milestone format. At plan time the planner queries by target name;
the retriever returns the canonical plan filtered by plan-prune,
dropping prefix steps already satisfied by the agent's inventory.
This is the \texttt{memory\_bank} component of combined retrieval
(\S\ref{app:mc_memory_ablation}); Optimus-1's AMEP module
\cite{li2024optimus} contributes visual context as
a separate joint signal. Retrieval is a target-name dictionary lookup
followed by the prune walk, with no LLM call and no embedding model. The
bank is serialized as a single JSON loaded once at startup and is
read-only during sweeps; failed plans flow into AMEP, not back into
the bank.

\subsection{Module mapping to desktop CUA}
\label{app:mc_mapping}
Table~\ref{tab:mc_mapping} shows the per-component correspondence between
the desktop and Minecraft instantiations.

\begin{table}[h]
\caption{Per-component correspondence between desktop and Minecraft
instantiations of the StructAgent framework. Minecraft actors and memory reuse
STEVE-1 \cite{lifshitz2023steve1}, JARVIS-1 \cite{wang2023jarvis}, and
Optimus-1 AMEP \cite{li2024optimus}.}
\label{tab:mc_mapping}
\small
\centering
\begin{tabular}{l p{0.32\linewidth} p{0.40\linewidth}}
\toprule
\tablehead
Component & Desktop CUA realization & Minecraft realization \\
\midrule
State (milestones)
& Instruction-derived requirement list.
& GPT-4o sub-task list of $(\textsc{verb}, \textsc{item}, n)$ tuples. \\
\midrule
Verifier (primary)
& Multi-stage screen / DOM / shell-output verifier.
& \texttt{task\_checker} (absolute inventory count) +
\texttt{SuccessMonitor} (sub-task) +
\texttt{VerificationRecord} (cumulative goal). Advances when the target inventory condition is verified. \\
\midrule
Failure attribution
& Routed intervention based on which verifier signal fired.
& \textsc{amep} failure metadata: failed sub-task, pending outcomes,
step count. Plan written to \texttt{plan/failed/}. \\
\midrule
Actor (action emission)
& GUI actor: emits action + coordinates in one call.
& \textsc{steve-1} (visuomotor) + \textsc{jarvis-1} craft helper +
smelt helper, sharing one dispatch interface. \\
\midrule
Memory (planning)
& Past trajectories indexed by task and DOM context.
& \textsc{amep} indexed by task, environment, and initial inventory. \\
\midrule
Memory (verification)
& Verifier examples and failure cases.
& Reflection memory + \texttt{memory\_bank} (recipe-derived canonical plans
from a 613-item / 848-recipe knowledge graph). \\
\midrule
Replanning
& Triggered when verifier rejects observed state.
& Triggered on missing-material detection inside craft helper. Plan-prune
applied once after retrieval against observed inventory. \\
\bottomrule
\end{tabular}
\end{table}

\subsection{Memory ablation: combined vs.\ sequential retrieval}
\label{app:mc_memory_ablation}
The Optimus-1 codebase \cite{li2024optimus} uses sequential retrieval in
\texttt{retrieve\_plan}:
\textsc{amep} is consulted first, and \texttt{memory\_bank} is consulted only
on \textsc{amep} miss. In a fresh-deployment regime, before tier-specific
trajectories have accumulated in \textsc{amep}, this systematically returns
a near-tier neighbor. With seven Stone-tier successes in \textsc{amep} and no
Iron-tier successes, every Iron-tier query under sequential retrieval receives
a Stone-tier nearest neighbor; the canonical Iron plan present in
\texttt{memory\_bank} is never consulted. The planner extrapolates the
Stone-tier example onto the Iron-tier task and truncates the plan at stone-tier
items. We observe this deterministically: on Iron task 2 under sequential
retrieval, the LLM produces a plan ending in \texttt{craft wooden\_axe} at
$\approx 1{,}820$ steps.

Our combined retrieval inverts the policy: \texttt{memory\_bank} is the
authoritative source of plan structure, \textsc{amep} supplements with the
most relevant recent visual context, and both are injected into the planner
prompt jointly. The same iron task under combined retrieval generates the
canonical 11-step plan and the agent reaches the target. On the same
eight-fix substrate, sequential retrieval produces $0/17$ Iron-tier passes
where combined retrieval produces $11/17$. Beyond pass rate, plan length is
qualitatively different: sequential retrieval yields 7--8 sub-task plans
truncated at wood or stone tier, combined retrieval yields the full 11-step
canonical chain.

\subsection{Failure modes}
\label{app:mc_failures}
Five distinct failure modes account for all observed Minecraft sweep failures.

\begin{table}[h]
\caption{Failure modes observed across the Wooden--Redstone sweeps.}
\label{tab:mc_failures}
\small
\centering
\begin{tabular}{p{0.20\linewidth} p{0.35\linewidth} p{0.35\linewidth}}
\toprule
\tablehead
Failure mode & Signature & Root cause \\
\midrule
Multi-cell craft retry storm
& 100+ verify-retry log lines on a single craft step; budget exhausted with
target absent from inventory.
& \textsc{jarvis-1} craft helper is GUI-blind: cursor moves computed via
camera-angle pixel math; sub-pixel drift makes placement clicks miss cells. \\
\midrule
Placed-block loss
& Agent crafts and opens a \texttt{crafting\_table}, walks away during
\textsc{steve-1} inference; the placed table is no longer reachable.
& \texttt{return\_crafting\_table()} exists upstream but its assertion is
commented out; silently fails when the policy moves the agent. \\
\midrule
Planner truncation
& LLM-generated plan ends at a wood- or stone-tier item when the target is
iron-tier. Deterministic on affected tasks.
& Without iron-tier trajectories in \textsc{amep}, sequential retrieval
returns a stone-tier nearest neighbor. Mitigated by combined retrieval
(\S\ref{app:mc_memory_ablation}). \\
\midrule
Ingredient over-ask
& Plan-prune drops every step except mining cobblestone; preload supplies 10,
plan asks for 11.
& \texttt{memory\_bank} canonical chains use worst-case ingredient counts;
the LLM does not normalize to recipe minimum. \\
\midrule
\textsc{steve-1} spatial blindness
& Explore-to-find reflection fires; thousands of steps with no resource gain.
& \textsc{steve-1} \cite{lifshitz2023steve1} uses Transformer-XL
\cite{dai2019transformerxl} with $\approx 6$\,s frame memory; text conditioning
decays; similar spatial-memory limits are discussed in MrSteve
\cite{park2025mrsteve}. \\
\bottomrule
\end{tabular}
\end{table}

The dominant failure on Iron and above is the multi-cell craft retry storm:
the agent reaches the final craft step at or near the budget cap, with
\textsc{SuccessMonitor}~=~1 on every prior sub-task but the target item
absent from the final inventory. This is orthogonal to the framework
contribution; replacing the GUI-blind craft helper with a pixel-aware GUI
verifier or a direct-API actor would resolve the failure.

\subsection{Case study: \texttt{wooden\_pickaxe} end-to-end}
\label{app:mc_case}
\begin{figure}[t]
\centering
\includegraphics[width=\linewidth]{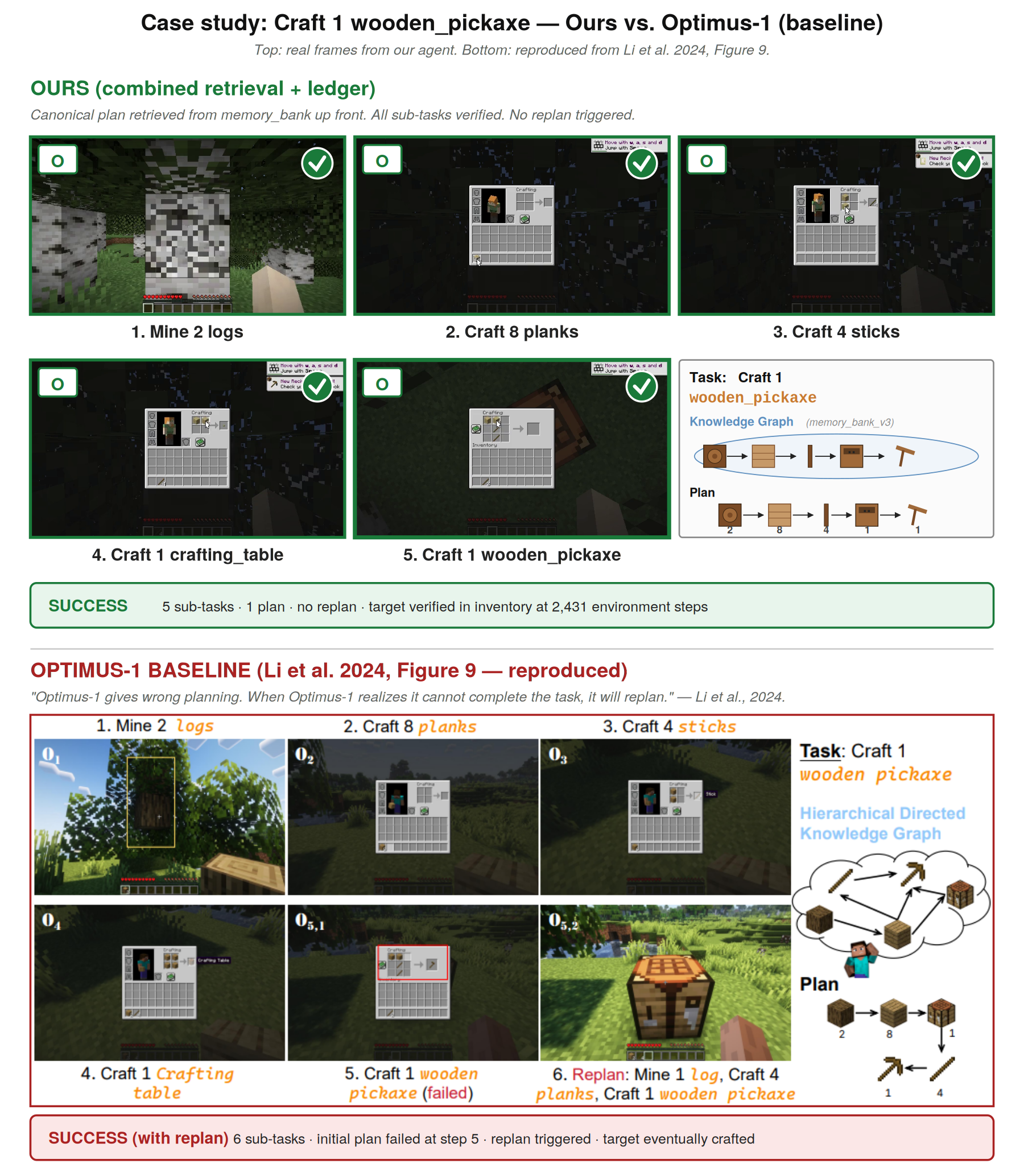}
\caption{Case study on \texttt{wooden\_pickaxe}. \textbf{Left:} our agent
under combined retrieval + verification record. Real frames from the recorded trajectory;
the canonical plan from \texttt{memory\_bank\_v3} occupies slot~6
(Task / Knowledge Graph / Plan). All five sub-tasks verify; no replan
triggered; target produced at 2{,}431 environment steps.
\textbf{Right:} Optimus-1 baseline, reproduced from
\citet{li2024optimus}~Fig.~9; the planner proposes an under-resourced plan
that fails at step 5 and recovers only via an in-flight replan.}
\label{fig:mc_case_study}
\end{figure}

Figure~\ref{fig:mc_case_study} contrasts our agent's execution of
\texttt{wooden\_pickaxe} against the published Optimus-1 trajectory for the
same task (\citet{li2024optimus}, Fig.~9). Under our combined retrieval,
\texttt{memory\_bank\_v3} supplies the canonical 5-step recipe chain
\texttt{logs}\,$\to$\,\texttt{planks}\,$\to$\,\texttt{sticks}\,$\to$\,%
\texttt{crafting\_table}\,$\to$\,\texttt{wooden\_pickaxe} up front, and the
agent completes all five sub-tasks in 2{,}431 environment steps with no
replan. The Optimus-1 baseline reaches the same first four sub-tasks but
proposes an under-resourced plan; the \texttt{wooden\_pickaxe} craft fails
at step 5 because the planner does not account for the planks consumed by
the prior \texttt{crafting\_table} step, and the agent must trigger an
in-flight replan (\texttt{mine 1 log}, \texttt{craft 4 planks},
\texttt{craft wooden\_pickaxe}) to recover.

The trace illustrates three properties relevant to the framework claim.
First, the state drives execution: each sub-task is dispatched, verified
by inventory delta, and only then marked complete. Second, the multi-stage
verifier catches the agent's actual progress: per-cell verify-retry recovers
from cursor misses inside the craft helper, \textsc{SuccessMonitor} confirms
inventory delta per sub-task, and the \texttt{VerificationRecord} declares
completion only when the target appears. Third, role separation is
preserved across the visuomotor, inventory-slot, and placed-block actors.
The replan loop visible in the baseline's bottom-right cell is the
observable signature of \textit{retrieval policy} as a design choice:
sequential retrieval surfaces a near-tier example whose ingredient counts
the LLM extrapolates inaccurately, while combined retrieval surfaces the
recipe-grounded canonical chain.

\section{Reproducibility}
\label{app:reproducibility}

\paragraph{Environment and model services.}
All desktop experiments use the OSWorld execution environment described in
Section~\ref{sec:setup}, with screenshot observations, a fixed
1920$\times$1080 display, and a 100-step interaction budget. Model services are
specified through runtime aliases for the planner/actor, verifier, and DONE
auditor. By default, the verifier follows the main model unless a separate
verifier model is configured, and the DONE auditor follows the verifier model.
This keeps model routing explicit in the saved configuration instead of being
hidden in prompt code.

\paragraph{Task splits and runtime configuration.}
OSWorld results use the OSWorld-Verified split described in
Section~\ref{sec:setup}. Mind2Web results use the protocol described in
Appendix~\ref{app:web_details}. The Minecraft study follows the task setting
described in Appendix~\ref{app:minecraft}. Structured memory is constructed
from the sources and indexing pipeline described in
Appendix~\ref{app:memory_tools}. Runtime configuration records the task split,
model aliases, step budget, screen size, memory-bank version, memory retrieval
depth, and enabled verifier/auditor modules.

\paragraph{Logged artifacts.}
Each run writes a result directory containing both final outcomes and
intermediate evidence. The top-level \texttt{args.json} records runtime
configuration. Each task directory stores the final task result, screenshots,
action traces, and module-specific debug files when the corresponding module is
enabled. Table~\ref{tab:repro_artifacts} summarizes the main artifacts used for
auditing.

\begin{table}[t]
  \centering\scriptsize
  \caption{Reproducibility artifacts saved by the runner.}
  \label{tab:repro_artifacts}
  \begin{tabularx}{\linewidth}{p{0.24\linewidth} p{0.24\linewidth} Y}
    \toprule
    \tablehead Artifact & Location & Purpose \\
    \midrule
    Runtime configuration & \texttt{args.json} & Records model aliases,
    task split, screen size, step budget, memory settings, and verifier/auditor
    flags. \\
    Task outcome & \texttt{result.txt} and retry summary files & Stores final
    success or failure status for aggregation. \\
    Verification trace & \texttt{boundary\_verify/} & Stores milestone
    verification decisions, evidence, and rejected or confirmed state updates. \\
    DONE audit trace & \texttt{done\_auditor/} & Stores final-completion audit
    decisions and supporting evidence when the auditor is enabled. \\
    Memory retrieval trace & \texttt{planner\_memory\_debug/} & Stores retrieved
    structured memories used by the planner. \\
    Failure diagnosis trace & \texttt{stuck\_diagnosis/} & Stores failure
    attribution and suggested interventions when diagnosis is enabled. \\
    \bottomrule
  \end{tabularx}
\end{table}

\paragraph{Scoring and aggregation.}
OSWorld scores are aggregated by the verified benchmark groups. Mind2Web results
use the WebJudge-style evaluation protocol described in
Appendix~\ref{app:web_details}. Minecraft metrics follow
Appendix~\ref{app:minecraft}. Aggregation is performed from saved result folders
rather than from live agent decisions, so the reported scores can be traced back
to the stored task outcomes and verification evidence.

\paragraph{Determinism and auditability.}
We fix task splits, display resolution, step budget, and runtime flags for each
controlled comparison. GUI execution and remote or distributed VLM serving are
not bit-exact across all reruns, so the release emphasizes matched
configuration and auditability: each reported run should include the saved
runtime configuration, task-level outcomes, screenshots, action traces, and
verification logs needed to inspect the result.

\clearpage

\section{Prompts and Structured Output Contracts}
\label{app:prompts}

This appendix reports the prompt templates and parsed contracts used by the
LLM-facing modules. The full runtime prompts are composed from a static role
prompt plus dynamic blocks, including the task, screenshots, accessibility
tree, state, memory retrieval, verifier traces, and domain action catalogs. The
excerpts below are copied from the implementation, with long catalogs and
examples removed only where explicitly marked.

\begin{table}[h]
  \centering\scriptsize
  \caption{LLM-facing prompt templates and contracts.}
  \label{tab:prompt_contracts}
  \begin{tabularx}{\linewidth}{p{0.16\linewidth} p{0.20\linewidth} p{0.23\linewidth} Y}
    \toprule
    \tablehead Module & Trigger & Output contract & Consumed by \\
    \midrule
    Verification Requirements Initializer & Task start & JSON verification record seed with
    \texttt{initial\_context}, \texttt{slots}, and
    \texttt{required\_outcomes}. & Seeds the \texttt{VerificationRecord} and
    the first milestone verification specs. \\
    Planner & Initial planning, progress check, and replan turns &
    XML-like tags for observation, subgoal assessment, decision, plan, and
    actor hint. & Updates the subgoal queue, strategy record, and the next
    actor instruction. \\
    Actor & Low-level execution turn & JSON action array, normalized into
    UI-TARS-style \cite{qin2025uitars} \texttt{Thought}/\texttt{Action} lines.
    & Compiled into
    GUI, host, or structured document actions. \\
    Low-level verifier & After an actor turn within an execution loop & One
    JSON object with \texttt{status}, evidence, and optional actor feedback. &
    Decides whether actor control should continue, retry, or return to the
    planner. \\
    Milestone verifier & Actor-loop exit and planner DONE boundary & JSON array
    of per-outcome judgments, optionally with one read-only probe. & Updates
    outcome states and proof traces in the verification record. \\
    DONE auditor & After the structured DONE gate accepts a planner DONE claim &
    Structured text ending with \texttt{VERDICT: PASS|FAIL|PARTIAL}. & Final
    adversarial check before accepting task completion. \\
    Failure attribution & Repeated stuck or rejected progress & JSON diagnosis
    with role attribution and intervention fields. & Routed to planner replan,
    actor redo, verifier override, environment intervention, or fallback. \\
    BBON judge & Section~\ref{sec:structured_evidence_analysis} analysis only &
    Per-candidate verdict lines plus \texttt{<choice>} and \texttt{<cite>}. &
    Selects one rollout among multiple candidates for test-time analysis. \\
    \bottomrule
  \end{tabularx}
\end{table}

\paragraph{Verification Requirement Initializer.}
At task start, the initializer receives the user instruction, the initial
screenshot, the accessibility tree, and environment-derived context. It returns
a JSON seed for the verification record. The prompt first asks the model to separate durable
milestones from transient waypoints:

\begin{promptlisting}[title={Verification Requirement initializer prompt excerpt}]
You initialize a structured progress verification record for a GUI
automation agent. The agent runs inside an OSWorld benchmark VM.

You see the user's task instruction and the agent's initial screen.
Your job is to read them and output:

  1. initial_context -- what the agent is starting with (app, active URL,
     window title, visible open tabs). Read these off the screen and the
     provided a11y tree. If you cannot read a field, leave it null.

  2. required_outcomes -- a small ORDERED chain of GATING MILESTONES that
     the agent must pass through to complete the task. NOT a checklist
     of every screen interaction; the planner breaks down the
     micro-steps.

Test: "If this state regressed, would the AGENT need to redo
OTHER work too -- i.e., does it CASCADE downstream?"

  YES -> MILESTONE   (include as outcome)
  NO  -> WAYPOINT    (DROP -- let the planner handle as subgoal)
\end{promptlisting}

The abridged schema below matches the implementation contract; omitted fields
are optional fields of the same \texttt{verify} object.

\begin{promptlisting}[title={Verification Requirements output contract}]
{
  "initial_context": {
    "open_tabs": ["<url>", ...],
    "active_url": "<url or null>",
    "active_app": "<app name or null>",
    "window_title": "<title or null>"
  },
  "value_provenance": [
    {"value": "...", "status": "KNOWN|UNKNOWN", "why": "..."}
  ],
  "slots": [
    {"name": "...", "type": "url|text|path|numeric|cell_ref",
     "description": "...", "source_hint": "..."}
  ],
  "required_outcomes": [
    {
      "id": "<snake_case>",
      "description": "<one sentence>",
      "evidence_hint": "<concrete observable>",
      "depends_on": ["<other_outcome_id>", ...],
      "verify": {
        "kind": "<one verify kind>",
        "file_path": "...",
        "patterns": ["<regex>", ...],
        "url_pattern": "...",
        "tag": "...",
        "text_contains": ["...", ...],
        "state_contains": ["...", ...],
        "visual_must_hold": ["...", ...],
        "command": ["<argv>", ...],
        "expected_substring": ["...", ...],
        "expected_exit_code": 0,
        "calc_checks": [{"op": "<op>", "...": "..."}],
        "impress_checks": [{"op": "<op>", "...": "..."}],
        "writer_checks": [{"op": "<op>", "...": "..."}]
      }
    }
  ]
}
\end{promptlisting}

\paragraph{Planner.}
The planner uses separate prompt builders for initial planning, progress
checking, and forced replanning, but the recurrent parser reads the same core
tags. The planner first describes observations and assesses the latest subgoal,
then emits a decision. A \texttt{REPLAN} decision may replace the queue with a
new \texttt{plan}; \texttt{CONTINUE} keeps the current queue; \texttt{DONE}
enters the structured DONE gate.

\begin{promptlisting}[title={Planner prompt excerpt}]
Output format -- emit these tags in EXACTLY this order:

<observations>
  <visible>CURRENT SCREEN -- the app / window / page in focus
  ... plus the INTERACTIVE elements you can act on right now.</visible>
  <unexplored>Content the task may need that is NOT in the
  current view.</unexplored>
</observations>

<last_subgoal_assessment>
  <evidence>One short sentence citing CONCRETE evidence that
  the previous subgoal's expected_post_state is -- or is not --
  satisfied.</evidence>
  <subgoal_anchor_check>...</subgoal_anchor_check>
  <contradiction_check>...</contradiction_check>
  <done>YES or NO</done>
</last_subgoal_assessment>

<Bottleneck>...</Bottleneck>
<decision>DONE|CONTINUE|REPLAN|INFEASIBLE</decision>
<plan>
  <strategy>...</strategy>
  <step>
    <text>...</text>
    <expected_post_state>...</expected_post_state>
  </step>
</plan>
<actor_hint>...</actor_hint>
\end{promptlisting}

The planner prompt also binds decisions to runtime effects. In the live prompt,
\texttt{DONE} requires every required outcome to be verified, the remaining
subgoal list to be empty, and no contradiction between verification record state and visible
state. This is what turns planner text into controlled state advancement rather
than a free-form completion claim.

\paragraph{Actor.}
The actor decomposes the current subgoal into a JSON array. Its base prompt and
the inline grounding directive are appended together at runtime:

\begin{promptlisting}[title={Actor prompt excerpt}]
You are a UI action decomposer. Given the current screenshot and a
subgoal, emit a JSON array of low-level actions to execute in order.
You output the on-screen pixel location of each element YOURSELF ...

STRICT FORMAT RULES -- violate any and downstream parsing fails:
- Respond with EXACTLY ONE JSON array. No prose before or after.
  No markdown fences (no ```json).
- Use double quotes, not single quotes, for every JSON string.
- Every "target" / "start_target" / "end_target" / "anchor_target"
  MUST be a description that pinpoints exactly one element visible in
  the screenshot ... you then give its coordinates yourself.
- GROUNDING-VISIBILITY RULE: every target ... is grounded against the
  CURRENT screenshot -- the one attached to this prompt -- not against
  some imagined future screenshot.
\end{promptlisting}

\begin{promptlisting}[title={Inline grounding directive}]
[INLINE GROUNDING MODE]
For EVERY action that targets a UI element, in ADDITION to the
existing "target" text field you MUST include the element's location
read directly from the screenshot, as coordinates normalized to a
0-1000 grid:
  - click / left_double / right_single: add "point": [x, y]
  - drag: add "start_point": [x, y] and "end_point": [x, y]
  - scroll: add "anchor_point": [x, y]
[omitted: non-spatial action list]
Example: {"action": "click", "target": "the Submit button",
"point": [840, 560]}
\end{promptlisting}

The parsed action contract used in the paper experiments is:

\begin{promptlisting}[title={Actor parsed action contract}]
[
  {"action": "click|left_double|right_single",
   "target": "<visible element>", "point": [x, y]},
  {"action": "drag",
   "start_target": "...", "end_target": "...",
   "start_point": [x, y], "end_point": [x, y]},
  {"action": "type",
   "text": "...", "wipe_then_type": false, "submit": false},
  {"action": "hotkey", "key": "..."},
  {"action": "scroll",
   "direction": "up|down|left|right",
   "anchor_target": "...", "anchor_point": [x, y]},
  {"action": "wait"},
  {"action": "impossible", "reason": "..."}
]
\end{promptlisting}

Host-side and structured-document actions such as \texttt{open\_app},
\texttt{navigate}, \texttt{extract\_info}, \texttt{cli\_run},
\texttt{edit\_json}, \texttt{calc\_*}, \texttt{impress\_*}, and
\texttt{writer\_*} are injected only when available for the current domain.

\paragraph{Low-level verifier.}
The low-level verifier judges only the current subgoal after one actor turn. Its
input includes the task, subgoal, expected post-state, actor response, executed
actions, before/after screenshots, up to three prior screenshots, and the recent
actor-action history. It does not update the verification record.

\begin{promptlisting}[title={Low-level verifier prompt excerpt}]
You are a strict low-level verifier for a computer-use agent.

Judge only the CURRENT SUBGOAL, not the whole task. Use the recent
actor-turn history for context, then judge the CURRENT actor turn by
comparing the CURRENT BEFORE and CURRENT AFTER screenshots with the
actor's last action.

Return:
- complete: the current subgoal is visibly satisfied now.
- continue: the action made useful progress, but the subgoal is not complete.
- no_effect: the action appears to have no useful visible effect.
- off_track: the agent moved to the wrong app/page/dialog or changed
  the wrong thing.
- stuck: the screen is blocked, loading, errored, or needs higher-level
  replanning.

Be conservative. Do not mark complete unless the AFTER screenshot clearly
supports the expected post-state. Output JSON only.
\end{promptlisting}

\begin{promptlisting}[title={Low-level verifier output contract}]
{
  "status": "complete|continue|no_effect|off_track|stuck",
  "reason": "one short sentence",
  "evidence": "brief visible evidence from the AFTER screenshot",
  "feedback_to_actor": "specific next low-level hint, or null"
}
\end{promptlisting}

\paragraph{Milestone verifier.}
The milestone verifier runs over target outcomes at a boundary. It first judges
from observation and may request one read-only deterministic probe when the
success condition is latent. A valid probe overrides the visible judgment; an
invalid or inconclusive probe falls back to the observation judgment. Anything
other than an explicit \texttt{verified} keeps the outcome unverified.

\begin{promptlisting}[title={Milestone verifier prompt excerpt}]
You are the MILESTONE VERIFIER for a GUI computer-use agent. The planner
just declared a subgoal complete. For each listed milestone, decide
whether it is OBJECTIVELY achieved on the CURRENT screen -- reason from
evidence; never rubber-stamp the planner's claim.

THE TASK INSTRUCTION IS THE GROUND TRUTH -- NOT THE MILESTONE LABEL.

LATENT STATE -> PROBE. A screenshot cannot prove a file's saved content,
a document's model (cell values / fonts / shapes), or the active URL.
When a milestone hinges on latent state, request ONE read-only check
rather than guessing from pixels.

NO EVIDENCE -> NOT VERIFIED. If you cannot quote concrete current evidence
that EVERY constraint holds, the milestone is not verified.
\end{promptlisting}

\begin{promptlisting}[title={Milestone verifier output contract}]
[
  {
    "outcome_id": "<id>",
    "constraints": "<task-specific requirements>",
    "observation": "<current a11y/screen/doc evidence>",
    "obs_verdict": "verified|not_verified|uncertain",
    "reason": "...",
    "evidence": "...",
    "probe": null
  }
]
\end{promptlisting}

The allowed probe kinds are
\texttt{file\_grep}, \texttt{url\_match}, \texttt{a11y\_match},
\texttt{calc\_verify}, \texttt{impress\_verify}, \texttt{writer\_verify}, and
guarded read-only \texttt{shell\_command}. The available kinds are filtered by
domain before they are shown to the verifier.

\paragraph{DONE auditor.}
The DONE auditor is a final audit, not another JSON verifier. It receives a
snapshot with the task text, current accessibility tree, current URL, verifier
specs, accepted proof claims, perceiver focus block, and recent screenshots. It
must parse the task first, cross-check evidence, list gaps when present, and end
with a single verdict line.

\begin{promptlisting}[title={DONE auditor prompt excerpt}]
You are the Done-Auditor for an OSWorld GUI agent. The agent has just
claimed the task is DONE. A separate structured verifier has already
approved this claim against its own checklist. Your job is to
adversarially refute that approval by examining only the raw evidence
you receive -- the user's literal task text, the screenshots you see,
the accessibility tree, and the proof traces the structured verifier
already accepted.

You do NOT see the planner's chain of thought, the plan steps, the
current subgoal text, or any strategy notes. This is by design -- you
are a fresh pair of eyes.

TASK TEXT IS TRUTH; verifier_specs are ONE LLM's guess at success
written at task start. When the spec marks an outcome PASS but the
task text names a target ... trust the task text and FAIL.
\end{promptlisting}

\begin{promptlisting}[title={DONE auditor output contract}]
TASK PARSE:
  <one bullet per required deliverable>

EVIDENCE CROSS-CHECK:
  <one bullet per task requirement with evidence>

GAP ANALYSIS (if any):
  <one bullet per gap>

VERDICT: PASS|FAIL|PARTIAL
\end{promptlisting}

\paragraph{Failure attribution and intervention.}
Failure attribution is invoked when the agent repeatedly fails to make verified
progress. The schema forces the model to recall relevant memory, observe the
current state, critique divergence from memory, attribute the failure to a role,
and propose an intervention. The downstream router consumes the role and
role-specific payloads rather than treating the diagnosis as free text.

\begin{promptlisting}[title={Failure attribution prompt excerpt}]
You are a root-cause failure-attribution analyst for a GUI agent stuck
on a single goal. You produce ONE JSON object -- nothing else. No prose
before or after. No markdown headings. No "Phase 1" labels. Just the JSON.

JSON field requirements (fill in this exact order so the conclusion is
grounded in earlier evidence):

1. memory_recall.canonical_approach: one sentence summarising what
RETRIEVED MEMORY recommends for this goal.
2. memory_recall.memory_referenced: array of 1-4 memory entries you
actually relied on.
3. observation_summary: 2 sentences MAX -- what the agent did, what the
current state actually shows.
4. evidence_quotes: array of 2-4 verbatim strings from the snapshot ...
Every claim downstream MUST be anchored to one of these.
\end{promptlisting}

\begin{promptlisting}[title={Failure attribution output contract}]
{
  "memory_recall": {
    "canonical_approach": "...",
    "memory_referenced": [
      {"layer": "L1|L2|L3a|L3c", "id": "...",
       "score": 0.0, "why_relevant": "..."}
    ]
  },
  "observation_summary": "...",
  "evidence_quotes": ["...", "..."],
  "memory_critique": {
    "memory_followed": "true|false|partial|n/a",
    "divergence_description": "..."
  },
  "root_cause_summary": "...",
  "attributed_to_role": "planner|actor|verifier|env|unclear",
  "category": "tactical|strategic|info_gap|unclear",
  "tactical_subkind": "<one tactical subkind or null>",
  "missing_or_wrong": ["..."],
  "confidence": "high|medium|low",
  "next_action_hint": "...",
  "verifier_override_evidence": [],
  "env_recovery_kind": null,
  "env_recovery_params": {}
}
\end{promptlisting}

Valid \texttt{tactical\_subkind} values are \texttt{wrong\_target},
\texttt{target\_missing}, \texttt{target\_blocked},
\texttt{blocking\_state}, \texttt{focus\_lost}, and
\texttt{execution\_error}; otherwise the value is \texttt{null}.
The router maps this diagnosis to one of five intervention decisions:
\texttt{planner\_replan}, \texttt{actor\_redo}, \texttt{verifier\_override},
\texttt{env\_intervention}, or \texttt{fallback\_planner}. Safety gates keep
role-specific interventions from firing when the evidence is weak.

\paragraph{BBON judge.}
BBON is used only in the analysis section. The judge receives a shared success
rubric and several candidate behavior narratives with decisive screenshots. It
first emits one verdict line per candidate, then chooses a single candidate.

\begin{promptlisting}[title={BBON judge prompt excerpt}]
You are an exacting judge selecting which candidate ACTUALLY completed a
computer-use task. You get a SUCCESS RUBRIC (the criteria) and, per
candidate, a behavior storyboard: objective actions + observed evidence
+ screenshots at the decisive steps, and the FINAL STATE.

METHOD:
  - For EACH rubric outcome, decide which candidates satisfy it FROM THE
EVIDENCE -- weight the FINAL STATE most. Do NOT trust any 'AGENT
SELF-CLAIM' line; agents frequently believe they succeeded when they did not.
  - Watch for an outcome reached then BROKEN later ...
  - Then pick the SINGLE candidate that best satisfies ALL outcomes.
\end{promptlisting}

\begin{promptlisting}[title={BBON judge output contract}]
<verdict>A: DONE|NOT_DONE -- evidence for candidate A</verdict>
<verdict>B: DONE|NOT_DONE -- evidence for candidate B</verdict>
...
<choice>A</choice><cite>one sentence citing decisive evidence</cite>
\end{promptlisting}

\paragraph{Parsing and fallbacks.}
All contracts are parsed defensively. Planner tags tolerate missing optional
blocks. Actor JSON is retried once if unparseable, then falls back to
\texttt{<Impossible>}. Low-level verifier parse failures default to
\texttt{continue}. Milestone verifier failures keep outcomes unverified, and
malformed probes are ignored. DONE-auditor transport or parse failures map to
\texttt{PARTIAL}. BBON choices are retried on invalid \texttt{<choice>} output
before the analysis code falls back and marks the parse failure.

\end{document}